%% file: main.tex
\documentclass[11pt]{article}
\PassOptionsToPackage{table}{xcolor}
\usepackage[preprint]{acl}
\usepackage{times,latexsym}
\usepackage[T1]{fontenc}
\usepackage[utf8]{inputenc}
\usepackage{microtype}
\usepackage{graphicx}
\usepackage{booktabs,multirow,array,tabularx,makecell,threeparttable}
\usepackage{xcolor}
\usepackage{amsmath,amssymb}
\usepackage{enumitem,pifont,xspace}
\usepackage[ruled,vlined,linesnumbered]{algorithm2e}
\SetAlgoNlRelativeSize{0}
\usepackage{placeins}
\usepackage{needspace}
\usepackage[most]{tcolorbox}
\usepackage{listings}

\definecolor{GainOne}{RGB}{239,245,250}
\definecolor{GainTwo}{RGB}{224,235,246}
\definecolor{GainThree}{RGB}{207,223,240}

\definecolor{LossOne}{RGB}{249,239,236}
\definecolor{LossTwo}{RGB}{244,224,219}
\definecolor{LossThree}{RGB}{237,205,198}

\definecolor{BlockGray}{RGB}{246,246,246}

\newcommand{\gainone}[1]{\cellcolor{GainOne}\rule{0pt}{2.15ex}#1}
\newcommand{\gaintwo}[1]{\cellcolor{GainTwo}\rule{0pt}{2.15ex}#1}
\newcommand{\gainthree}[1]{\cellcolor{GainThree}\rule{0pt}{2.15ex}#1}

\newcommand{\lossone}[1]{\cellcolor{LossOne}\rule{0pt}{2.15ex}#1}
\newcommand{\losstwo}[1]{\cellcolor{LossTwo}\rule{0pt}{2.15ex}#1}

\newcommand{\blockrow}[1]{\rowcolor{BlockGray}\multicolumn{10}{@{}l}{\textbf{#1}}}

\newcommand{\method}{\textsc{RubricWorld}\xspace}

\definecolor{ACLLinkBlue}{HTML}{000099}
\hypersetup{colorlinks=true,linkcolor=ACLLinkBlue,citecolor=ACLLinkBlue,urlcolor=ACLLinkBlue}
\newcommand{\LinkSection}[1]{\hyperref[#1]{\textcolor{ACLLinkBlue}{Section~\ref*{#1}}}}
\newcommand{\LinkAppendix}[1]{\hyperref[#1]{\textcolor{ACLLinkBlue}{Appendix~\ref*{#1}}}}
\makeatletter
\newcommand{\AppendixContentsLine}[3]{%
  \begingroup
  \ifnum#1=1\relax
    \addvspace{4pt}\def\AppendixIndent{0em}\bfseries
  \else
    \def\AppendixIndent{1.5em}
  \fi
  \@dottedtocline{#1}{0em}{2.7em}%
    {\hspace*{\AppendixIndent}\hyperref[#2]{\textcolor{ACLLinkBlue}{\numberline{\ref*{#2}}#3}}}%
    {\hyperref[#2]{\textcolor{ACLLinkBlue}{\pageref*{#2}}}}%
  \endgroup}
\makeatother

\definecolor{promptbg}{HTML}{F7F7F7}
\definecolor{promptframe}{HTML}{D6D6D6}
\definecolor{prompttitle}{HTML}{333333}

\lstdefinestyle{promptstyle}{
  basicstyle=\ttfamily\footnotesize,
  breaklines=true,
  breakatwhitespace=false,
  columns=fullflexible,
  keepspaces=true,
  showstringspaces=false,
  tabsize=2,
  frame=none
}

\newtcblisting{promptbox}[2][]{
  enhanced,
  breakable,
  listing only,
  colback=promptbg,
  colframe=promptframe,
  coltitle=prompttitle,
  fonttitle=\bfseries\footnotesize,
  title={#2},
  arc=1.5mm,
  boxrule=0.4pt,
  left=1mm,
  right=1mm,
  top=1mm,
  bottom=1mm,
  listing options={style=promptstyle},
  #1
}

\title{\method: World Model-Guided Rubric Optimization for LLM Graders}

\author{%
  \textbf{Hongji Pu}\textsuperscript{1} \quad
  \textbf{Hang Li}\textsuperscript{2} \quad
  \textbf{Yucheng Chu}\textsuperscript{2} \quad
  \textbf{Jiliang Tang}\textsuperscript{2}%
  \thanks{Corresponding author.} \\
  \textsuperscript{1}University of Illinois Urbana-Champaign,
  Urbana, IL, USA \\
  \textsuperscript{2}Michigan State University,
  East Lansing, MI, USA \\
}

\begin{document}
\maketitle
\input{sections/0_abstract}
\input{sections/1_introduction}
\input{sections/2_related}

\input{sections/3_methodology}
\input{sections/4_experiment}
\input{sections/6_conclusion}

\bibliography{references}
\clearpage
\clearpage
\appendix
\input{sections/Appendix}
\end{document}

%% file: sections/0_abstract.tex
\begin{abstract}
Rubric optimization improves LLM grading by revising scoring
instructions while keeping the grader's parameters fixed.
Existing optimizers often judge edits by aggregate validation
performance, limiting the supervision they retain about which
errors an edit corrects or introduces.
Testing each candidate also requires rescoring validation answers,
which limits how many edits can be explored within a fixed budget.
We propose \method, a framework that learns from previous edit
evaluations to guide further search.
Its world model predicts an edit's correction and degradation rates
at score boundaries, conditioned on the current rubric state.
The training signal comes from paired grading results: scores for
the same answers before and after an edit.
Predictions select one candidate for direct evaluation, and held-out
confidence bounds determine whether it corrects enough target errors
while keeping new errors within prespecified limits.
Measured effects update the model even when the edit is rejected.
Across EIR (National), ASAP2.0, and ASAP-SAS with three frozen LLM
graders, \method achieves the highest score in 20 of 27 metric
comparisons.
On ASAP2.0, it raises \texttt{gpt-5-mini} accuracy from 18.20\%
with the original rubric to 46.70\%, exceeding the strongest
baseline, APO, by 22.60 percentage points.
\end{abstract}

%% file: sections/1_introduction.tex
\section{Introduction}
\label{sec:introduction}

Automatic scoring has evolved from essay graders based on hand-crafted
features to learned scoring systems
\citep{page1966imminence,attali2006erater,taghipour2016neural}.
Large language models (LLMs) can score open-ended answers using
natural-language rubrics, examples, and scoring criteria
\citep{mansour2024score,henkel2024makegrade,hashemi2024llmrubric,zhou2025graders}.
Vague or incomplete rules can cause grading errors.
Revising the rubric can improve grading without retraining the model
\citep{gradeopt,chu2026caro}, while keeping the changes available for
teacher review \citep{henkel2025humans}.
Testing each revision, however, consumes model calls and tokens.
Effective optimization therefore depends on how much the optimizer
learns from each evaluation.

Existing rubric and prompt optimizers generate revisions from grading
errors, textual critiques, or evaluation traces
\citep{pryzant2023apo,gradeopt,chu2026caro,agrawal2025gepa}.
They test revisions on labeled examples and use the results to guide
search. However, they do not explicitly learn from past edits to
predict which errors a new edit will correct or introduce under the
current rubric.
When candidate quality is summarized by validation accuracy,
corrections can outweigh and conceal newly introduced errors.
For example, raising the requirements for score~4 may correctly lower
over-scored answers from~4 to~3, but also lower answers that deserve~4.
The same net accuracy gain could reflect a few corrections or many
corrections offset by many new errors.
Accuracy alone gives limited guidance on how to preserve the
corrections while avoiding the new errors.

Finding a suitable revision is also costly.
Several edits may address the same problem, and testing each requires
the grader to rescore validation answers
\citep{yang2024opro,opsahlong2024mipro}.
This limits the number of candidates explored within a fixed budget.
Moreover, earlier revisions alter the rubric under which later edits
are evaluated: adding or strengthening one rule can change another
edit's effect \citep{sclar2024sensitivity}.
Reusing past evaluations therefore requires accounting for the
current rubric, not just the candidate text.

\begin{figure*}[t]
    \centering
    \includegraphics[width=\textwidth]{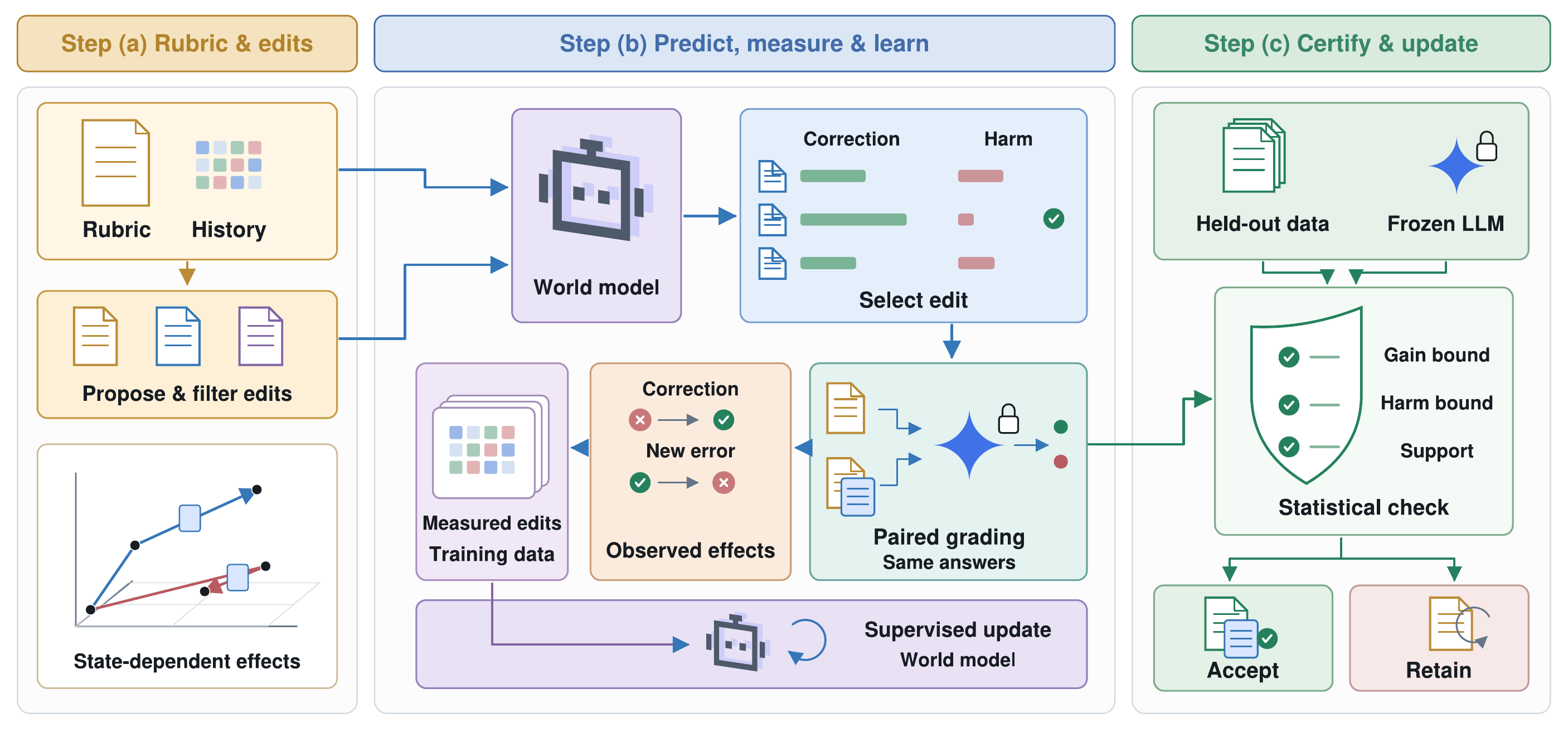}
    \caption{\textbf{Overview of \method.}
    (a) The current rubric and observed grading errors guide the
    generation of edits with a limited scope.
    (b) The world model predicts corrections and degradations for each
    candidate and guides the choice of an edit for evaluation.
    Paired grading decisions before and after the edit provide learning
    feedback, including for edits that are later rejected.
    (c) Confidence bounds on held-out answers determine whether the edit
    is accepted. The LLM grader's parameters remain fixed throughout.}
    \label{fig:rubricworld_pipeline}
\end{figure*}

\textbf{To address these problems, we propose \method, a world
model-guided framework for rubric optimization.}
Here, a world model predicts the grading effect of an edit from the
current rubric state and the edit itself.
This lets the optimizer reuse earlier evaluations while accounting for
the rubric changes made since those evaluations.
It can then screen proposed edits before spending grader calls on them.

We train the model on \emph{paired grading results}: the scores assigned
to the same answers before and after an edit, compared with their human
reference scores.
To describe what changed, we express each score as decisions at ordinal
score boundaries \citep{mccullagh1980ordinal,cao2020coral}.
For example, a score of~4 lies above the $1|2$, $2|3$, and $3|4$
boundaries.
At each boundary, the paired results reveal which incorrect decisions
became correct and which correct decisions became incorrect.
Together, these correction and degradation rates define the
\emph{edit effect}, including effects away from the edit's intended
boundary.
The world model learns these rates from the edit and its interaction
with the current rubric state.

In each round, \method uses the predicted effects to choose one
candidate for direct evaluation, reducing the number of candidates
that require rescoring answers.
Separate held-out answers determine whether to accept the edit:
confidence bounds must support sufficient target correction and
limited degradation at the boundaries chosen for certification.
The measured training-set effect is retained even when the edit is
rejected, so that unsuccessful attempts also inform later selections.

Our contributions are:
\begin{itemize}[topsep=2pt,itemsep=0pt,parsep=0pt,partopsep=0pt]
    \item We use paired grading results to measure an edit's
    corrections and degradations at each ordinal score boundary.
    \item We introduce \method, which learns to predict these
    effects under the current rubric and uses them to select candidates,
    with independent held-out checks for rubric updates.
    \item We evaluate grading quality, optimization cost, and
    accumulated errors on EIR (National), ASAP2.0, and ASAP-SAS
    with three frozen LLM graders.
\end{itemize}

%% file: sections/2_related.tex
\section{Related Work}
\label{sec:related}

\textbf{Rubric-Guided LLM Evaluation.}
G-Eval uses task-specific evaluation steps to guide LLM scoring, while
Prometheus learns fine-grained evaluation from supplied rubrics
\citep{liu2023geval,kim2023prometheus}.
LLM-Rubric further combines and calibrates criterion-level judgments for
more structured evaluation \citep{hashemi2024llmrubric}.
These methods primarily study how a given scoring rubric is applied or
calibrated, rather than how it should be revised repeatedly from
observed grading errors
\citep{kim2023prometheus,hashemi2024llmrubric}.

\textbf{Feedback-Driven Prompt and Rubric Optimization.}
APO and GEPA improve instructions using textual feedback or evaluation
traces, while GradeOpt derives grading guidelines from misgraded
answers \citep{pryzant2023apo,agrawal2025gepa,gradeopt}.
PromptAgent formulates prompt refinement as sequential planning, whereas
MIPRO learns a surrogate objective to guide candidate search
\citep{wang2023promptagent,opsahlong2024mipro}.
Recent optimizers also introduce preservation or degradation constraints
to reduce losses on previously correct behavior
\citep{xu2026steve,satheesha2026rlmopt}.
These approaches improve candidate generation, search, or update
selection, but do not explicitly learn the paired edit effect conditioned
on the current rubric state; this matters because earlier revisions
change the state under which later edits are applied
\citep{wang2023promptagent,opsahlong2024mipro}.

\textbf{World Model Learning.}
Recent world-model methods learn action-conditioned dynamics and use
predicted outcomes to guide subsequent planning or control
\citep{hansen2024tdmpc2,hafner2025dreamer}.
\method adopts this predictive view at the level of a single rubric
edit: rather than predicting only a scalar prompt score, it models
paired corrections and degradations across score boundaries under the
current rubric state. The model predicts the effect of one edit at a time. These predictions
guide candidate selection, and separate held-out evidence determines
whether to accept the edit.

%% file: sections/3_methodology.tex
\section{Methodology}
\label{sec:method}

\method is a framework for improving a frozen LLM grader by editing
its rubric. It has three steps: generate candidate edits, predict their
effects, and check the selected edit before accepting it
(Figure~\ref{fig:rubricworld_pipeline}).
Each round, the generator proposes edits for a group of observed
grading errors. The world model predicts which errors each edit will
correct or introduce under the current rubric and selects one candidate
for direct evaluation. Independent held-out answers determine whether
to accept it. Its measured training-set effect updates the world model
even if the rubric is retained.

\subsection{Measuring an Edit's Effect}
\label{sec:problem_formulation}
\label{sec:edit_effect}

An edit can correct some decisions while making others wrong.
We measure these two effects separately at each score boundary.
Let $f$ be the frozen LLM grader, $R_t$ the rubric at round $t$,
and $F_f(x;R)$ its score for answer $x$ under rubric $R$.
For answer $x_i$ with human score $y_i$, the current prediction is
$\hat y_i^{(t)}=F_f(x_i;R_t)$.
Applying edit $p$ gives the temporary rubric $R_t\oplus p$ and score
$\hat y_i^p=F_f(x_i;R_t\oplus p)$; $\oplus$ denotes the text modification.
Grading the same answers under both rubrics gives the paired observations.

For $K=y_{\max}-y_{\min}+1$ consecutive integer scores, the set
$\mathcal B=\{y_{\min},\ldots,y_{\max}-1\}$ contains the $K-1$
boundaries, where $b$ separates scores $b$ and $b+1$
\citep{mccullagh1980ordinal,cao2020coral}.
At each boundary, the question is whether an answer deserves a score
above $b$. The human-score label is $z_{i,b}=\mathbb I(y_i>b)$,
and the grader's before- and after-edit decisions are
$\hat z_{i,b}^{(t)}=\mathbb I(\hat y_i^{(t)}>b)$ and
$\hat z_{i,b}^{p}=\mathbb I(\hat y_i^p>b)$, where $\mathbb I$ is the
indicator function.

Let $\mathcal F_{t,b}$ contain the initially incorrect decisions in
the evaluated batch, and $\mathcal A_{t,b}$ the initially correct ones.
Correction is the fraction of $\mathcal F_{t,b}$ made correct by the
edit; degradation is the fraction of $\mathcal A_{t,b}$ made incorrect:
\begin{align}
C_{t,b}(p)
&=\frac{\sum_{i\in\mathcal F_{t,b}}
    \mathbb I(\hat z_{i,b}^{p}=z_{i,b})}{|\mathcal F_{t,b}|},
\label{eq:correction_rate}\\
D_{t,b}(p)
&=\frac{\sum_{i\in\mathcal A_{t,b}}
    \mathbb I(\hat z_{i,b}^{p}\neq z_{i,b})}{|\mathcal A_{t,b}|}.
\label{eq:degradation_rate}
\end{align}
For example, correcting 6 of 20 initially wrong decisions gives
$C=0.30$, while making 2 of 80 initially correct decisions wrong
gives $D=0.025$. A net accuracy change would combine these effects.

The edit-effect vector $\mathbf e_t(p)=[\mathbf C_t(p);\mathbf D_t(p)]$
collects these rates in increasing boundary order and has
$d_e=2(K-1)$ coordinates.
It records effects at both the target boundary and other boundaries.
If a denominator is empty, its rate is undefined and excluded from
training by a binary mask $\mathbf m_t(p)$; it is not recorded as zero.
We use only answer records with complete human, before-edit, and
after-edit scores (\LinkAppendix{app:response_masks}).

Let $\mathcal P_t$ be the candidate set, $b_t^\star$ the target boundary,
and $\epsilon_b$ the maximum tolerated degradation at boundary $b$.
The round-level objective is
\begin{equation}
\label{eq:rubric_objective}
\begin{aligned}
\underset{p\in\mathcal P_t}{\operatorname{maximize}}
&\quad C_{t,b_t^\star}(p)\\
\text{subject to}
&\quad D_{t,b}(p)\leq\epsilon_b,\quad b\in\mathcal B.
\end{aligned}
\end{equation}
This rewards target corrections while limiting new errors separately
at each boundary. It does not directly maximize exact-score accuracy
(ACC) or quadratic weighted kappa (QWK) \citep{cohen1968weighted}.

\subsection{Data Use and Candidate Generation}
\label{sec:data_candidates}

\paragraph{Data requirements.}
Each example contains an item, an answer, and its human score.
\method requires three disjoint labeled sets:
$\mathcal D_{\mathrm{dyn,train}}$ for candidate generation,
paired-effect measurement, and world-model training;
$\mathcal D_{\mathrm{dyn,val}}$ for tuning; and
$\mathcal D_{\mathrm{cert}}$ for independent acceptance checks.
Certification requires a fresh, disjoint subset
$\mathcal D_{\mathrm{cert}}^{(t)}$ at each round to prevent adaptation
to certification answers \citep{dwork2015holdout}.
$\mathcal D_{\mathrm{test}}$ is reserved for final evaluation.
Section~\ref{sec:baseline_setup} describes their construction.

\paragraph{Error groups and candidates.}
At each boundary, over-scoring means that the grader places an answer
above the boundary when its human score does not; under-scoring is the
reverse. Thus there are at most $2(K-1)$ boundary--direction groups.
We further separate errors present under the initial rubric from errors
introduced by accepted edits, giving at most $4(K-1)$ groups.
An answer can occur in several groups if its score crosses several
boundaries. Empty groups are omitted.

The routing rule selects one nonempty group from the training errors,
fixing the target boundary $b_t^\star$ and correction direction before
candidate generation (\LinkAppendix{app:error_groups}).
Given this group, the current rubric, and relevant correctly graded
training answers, the generator proposes $M=5$ local edits.
Each edit tightens, loosens, or clarifies the chosen score distinction.
We remove duplicate rules, contradictions, unrelated criteria, and
rules tied to a particular answer's wording.
The remaining candidates form $\mathcal P_t$; each retains its target,
direction, supporting examples, and proposed text change
(\LinkAppendix{app:scoped_candidate_generation}).
If no training errors remain, $\mathcal P_t=\varnothing$.
\LinkAppendix{app:contents-boundary-targeted-patch-schema} gives an
illustrative edit.

\subsection{Predicting Edit Effects}
\label{sec:effect_model}

Measuring every candidate's effects requires repeated grader calls.
The world model predicts those effects first, so only the selected
candidate needs direct evaluation.
Its input is the edit $p$ and state $s_t=(R_t,\mathcal H_t,\mathbf v_t)$.
Here, $\mathcal H_t$ stores previously measured edit effects, and
$\mathbf v_t$ summarizes the grader's current scores and boundary
correctness on training answers, together with answer features.
These inputs describe what has already been tried and which errors
remain. Only $R_t$ is supplied to the grader as its scoring rubric;
the history and summary are inputs to the world model.

\paragraph{Text embeddings.}
A general-purpose text encoder embeds the proposed edit together
with its target boundary and correction direction, giving
$a_p=\phi(p)\in\mathbb R^{d_a}$.
The state encoder combines an embedding of the rubric text with the
measured history and training-answer summary to produce
$h_t=\psi(s_t)\in\mathbb R^{d_h}$.
Text embeddings represent what the rules say; paired grading
observations teach the predictor how those rules affect the grader.
The encoders operate separately from the grader and require no access
to its weights, so the same interface supports API-based graders.
Optional score-probability features are used only when available;
details are in \LinkAppendix{app:encoders}.

\paragraph{Edit and state interaction.}
With trainable parameters $\theta$, the predictor learns an edit's
general effect and an adjustment for the current rubric state:
\begin{equation}
\label{eq:effect_logits}
\begin{aligned}
\boldsymbol\eta_\theta(s_t,p)
&=W_a a_p+c+U[(A a_p)\odot(B h_t)],\\
\boldsymbol\mu_\theta(s_t,p)
&=\sigma(\boldsymbol\eta_\theta(s_t,p)).
\end{aligned}
\end{equation}
The learned map $W_a$ predicts from the edit alone, with bias $c$.
The learned maps $A$ and $B$ project the edit and state to the same
width; element-wise multiplication $\odot$ combines them, and $U$
maps the result to an adjustment for each correction or degradation
rate \citep{rendle2010fm}.
The sigmoid $\sigma$ keeps predicted rates in $[0,1]$.
This interaction allows the same edit to receive different predictions
under different rubrics. Matrix dimensions are given in
\LinkAppendix{app:factorized_interaction}.

\paragraph{Learning from measured edits.}
Each record $(s,p,\mathbf e,\mathbf m)$ in $\mathcal T_t$ contains
the state, edit, measured rates, and validity mask. We minimize
\begin{equation}
\label{eq:effect_loss}
\mathcal L(\theta)=
\sum_{(s,p,\mathbf e,\mathbf m)\in\mathcal T_t}
\sum_{\substack{1\leq k\leq d_e\\m_k=1}}
\ell\!\left(\mu_{\theta,k}(s,p),e_k\right),
\end{equation}
where $\ell$ is the scalar regression loss and $k$ indexes a rate.
The mask excludes undefined rates. Initial records $\mathcal T_0$
come from edits evaluated on training answers before model-guided
selection. We fit five independently seeded models and update them
as new edits are measured (\LinkAppendix{app:response_model_details}).

\subsection{Selecting and Evaluating a Candidate}
\label{sec:selection}

The five models provide both a mean prediction and their disagreement
about each candidate \citep{lakshminarayanan2017deepensembles}.
For candidate $p_t^j$ and boundary $b$, let $\mu^C_{t,j,b}$ and
$\mu^D_{t,j,b}$ be the mean predicted correction and degradation rates,
and $\sigma^C_{t,j,b}$ and $\sigma^D_{t,j,b}$ their sample standard
deviations across models.
We penalize disagreement to avoid choosing an edit only because of
an uncertain favorable prediction. With penalty weight $\beta\geq0$,
\begin{equation}
\label{eq:predicted_scores}
\begin{aligned}
\widetilde C_{t,j}
&=\mu^C_{t,j,b_t^\star}-\beta\sigma^C_{t,j,b_t^\star},\\
\widetilde D_{t,j,b}
&=\mu^D_{t,j,b}+\beta\sigma^D_{t,j,b}.
\end{aligned}
\end{equation}
These are screening scores, not confidence bounds with a coverage
guarantee \citep{guo2017calibration}.
We select the candidate with the largest penalized target correction
among those within the predicted degradation limits:
\begin{equation}
\label{eq:candidate_selection}
\begin{aligned}
p_t^\star\in
&\arg\max_{p_t^j\in\mathcal P_t}\widetilde C_{t,j}\\
\text{subject to}
&\quad\widetilde D_{t,j,b}\leq\epsilon_b,\quad b\in\mathcal B.
\end{aligned}
\end{equation}
If none qualifies, the round leaves the rubric and model unchanged.
Otherwise, paired grading on $\mathcal D_{\mathrm{dyn,train}}$ measures
the selected edit's effect.
We add this observation to the training records $\mathcal T_t$ and
the state history $\mathcal H_t$, then update all five models.
The training records support fitting across states; the history tells
the predictor what has already been tried in the current run.
Rejected edits remain in both: their measured effects help the model
recognize similar unsuccessful candidates later.

\subsection{Checking Whether to Accept the Edit}
\label{sec:certification}

A favorable prediction is not enough to update the rubric.
We evaluate the selected edit on the unused answers in
$\mathcal D_{\mathrm{cert}}^{(t)}$ and require evidence that it corrects
enough target errors while introducing few new ones.
We call this check \emph{certification}.

The certification scope $\mathcal B_{\mathrm{cert}}\subseteq\mathcal B$
lists the boundaries for which the experiment requires this check.
It is fixed before candidate outcomes are inspected, so a boundary
cannot be dropped because an edit performed poorly there.
The target $b_t^\star$ must belong to this scope.
All boundaries still enter predictive screening, but statistical
claims cover only the prespecified scope.
The required minimum correction is $\gamma=0.80$ and the maximum
degradation is $\epsilon_b=0.08$.

Each initially wrong decision either becomes correct or remains
wrong; each initially correct decision either becomes wrong or
remains correct. Under independent binomial sampling, these counts
support exact one-sided confidence bounds
\citep{clopper1934confidence}.
We use a lower bound for target correction, because it must be large
enough, and an upper bound for degradation, because it must stay small.
For per-edit error probability $\delta$, we divide the error budget
equally among the one correction bound and the
$|\mathcal B_{\mathrm{cert}}|$ degradation bounds.
This accounts for checking several requirements on the same edit.

Let $L^C_{t,b_t^\star}$ be the target-correction lower bound and
$U^D_{t,b}$ the degradation upper bound at boundary $b$, computed
from the corresponding initially wrong or correct decisions.
The edit is accepted only if
\begin{equation}
\label{eq:certification_rule}
\begin{aligned}
L_{t,b_t^\star}^{C}&\geq\gamma,\\
U_{t,b}^{D}&\leq\epsilon_b,\quad b\in\mathcal B_{\mathrm{cert}}.
\end{aligned}
\end{equation}
If a required sample is too small to establish its bound even in the
best case, the edit is marked unsupported. Otherwise, failing a bound
gives a rejection. Both outcomes retain the rubric.
The binomial formulas, sample requirements, and assumptions are in
\LinkAppendix{app:certification_details}.
The coverage statement applies to one edit; a joint guarantee across
rounds requires additional confidence allocation.

An accepted edit gives $R_{t+1}=R_t\oplus p_t^\star$; otherwise
$R_{t+1}=R_t$.
We then refresh the training-answer summary under the resulting rubric
to form the next state. After $T$ rounds, $R_T$ is used for test scoring.

%% file: sections/4_experiment.tex
\section{Experiments}
\label{sec:experiments}

\subsection{Experimental Design}
\label{sec:experimental_setup}

We evaluate final grading quality, optimization cost, the role of the
current rubric state, and the contributions of prediction, selection,
and certification. We also track grading errors over 100 editing rounds.

\input{sections/table_main_results}

\begin{figure*}[!t]
\centering
\includegraphics[width=0.955\textwidth]{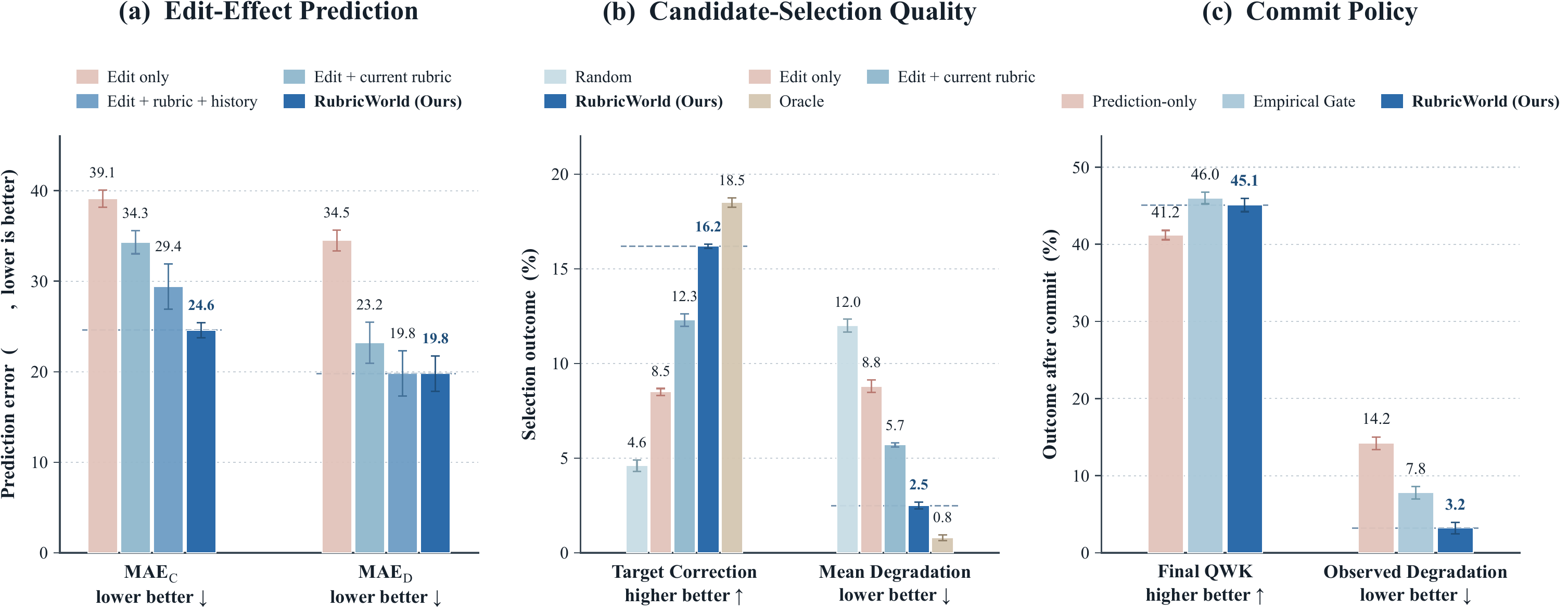}
\caption{\textbf{Prediction, selection, and certification on ASAP2.0 (\texttt{gpt-5-mini}).}
(a) Held-out-state MAE (pp); (b) selected-edit correction and degradation (\%);
(c) test QWK (0--100) and degradation (\%). Three-seed mean\,$\pm$\,SD;
variants in \LinkAppendix{app:prediction_variants}.}
\label{fig:ablation_triptych}
\end{figure*}

\noindent\textbf{Benchmarks.}
EIR (National) contains five private short-answer items, with
item-specific scores from 0 to 3 or 0 to 4
\citep{gradeopt}.
ASAP2.0 contains student essays scored from 1 to 6
\citep{learningagency2024aes2,crossley2025asap2}.
ASAP-SAS evaluates short answers; the reported Set~1 uses scores
from 0 to 3 \citep{kaggle2013asapsas,sakaguchi2015effective}.
The ordered score scales allow us to examine both exact scores
and decisions at each adjacent score boundary.
Counts and task-level splits are in \LinkAppendix{app:dataset_protocol}.

\noindent\textbf{Frozen LLM graders.}
We use Llama-3.1-8B-Instruct, an 8B-parameter instruction-tuned
language model \citep{grattafiori2024llama3}; QwQ-32B-Preview,
a 32B-parameter reasoning language model from Qwen
\citep{qwq2024preview}; and OpenAI's API-based language model
\texttt{gpt-5-mini} \citep{openai2026gpt5mini}.
Llama and QwQ use the 4-bit activation-aware weight-quantized (AWQ)
checkpoints in Table~\ref{tab:main_results} \citep{lin2024awq}.
Only the scoring rubric is updated.

\subsection{Methods, Protocol, and Measures}
\label{sec:baseline_setup}

\noindent\textbf{Comparison methods.}
Zero-shot uses the original rubric, and few-shot adds labeled
examples without changing it \citep{brown2020language}.
APO edits instructions using language-based feedback on mistakes
\citep{pryzant2023apo}; GradeOpt refines grading guidelines from
misgraded answers \citep{gradeopt}; and GEPA uses evaluated traces,
reflection, and a Pareto-based prompt search
\citep{agrawal2025gepa}.
All methods begin with the same rubric and frozen grader.

\noindent\textbf{Editing and evaluation.}
The main comparison uses five refinement rounds and three
random seeds. \method considers five candidate edits per round.
Before editing, we reserve test answers and then the certification
pool, remove cross-pool duplicates, and sample dynamics-validation
then dynamics-training answers, stratified by human score with fixed
seeds. Certification uses disjoint round-specific subsets
(\LinkAppendix{app:dataset_protocol}; \LinkAppendix{app:data_isolation}).
Predictor variants use the same edit embeddings.
Baselines use their native search procedures within the declared
refinement protocol; implementation details are in
\LinkAppendix{app:baseline_details}.

\noindent\textbf{What we measure.}
\label{sec:evaluation_metrics}
We report exact-score accuracy (ACC), quadratic weighted kappa
(QWK), and macro-F1 for final grading quality
\citep{cohen1968weighted,attali2006erater}.
For edit-effect prediction, we report errors in the predicted
correction and degradation rates ($\mathrm{MAE}_C$ and
$\mathrm{MAE}_D$) in percentage points (pp); comparisons use unseen
rubric states rather than random transition splits.
For selection, we measure the chosen edit's actual target
correction, mean degradation, and empirical feasibility rate (EFR).
For acceptance rules, we compare final QWK and observed degradation
under different acceptance rules. Metric definitions appear in
\LinkAppendix{app:metric_details}.

\noindent\textbf{Cost and long-horizon protocol.}
Optimization cost counts LLM tokens spent on proposing edits,
collecting warm-start and online observations, paired scoring, and
independent commit checks, excluding the shared final evaluation.
The cost sweep reports three-seed mean QWK versus optimization tokens
for \texttt{gpt-5-mini}.
\Needspace{4\baselineskip}
The 20-round continuation retains the five-round settings.
The 100-round plots use matched ASAP2.0 / \texttt{gpt-5-mini} runs,
tracking validation ACC and residual degradation separately.

\begin{figure}[!htbp]
\centering
\includegraphics[width=0.99\columnwidth]{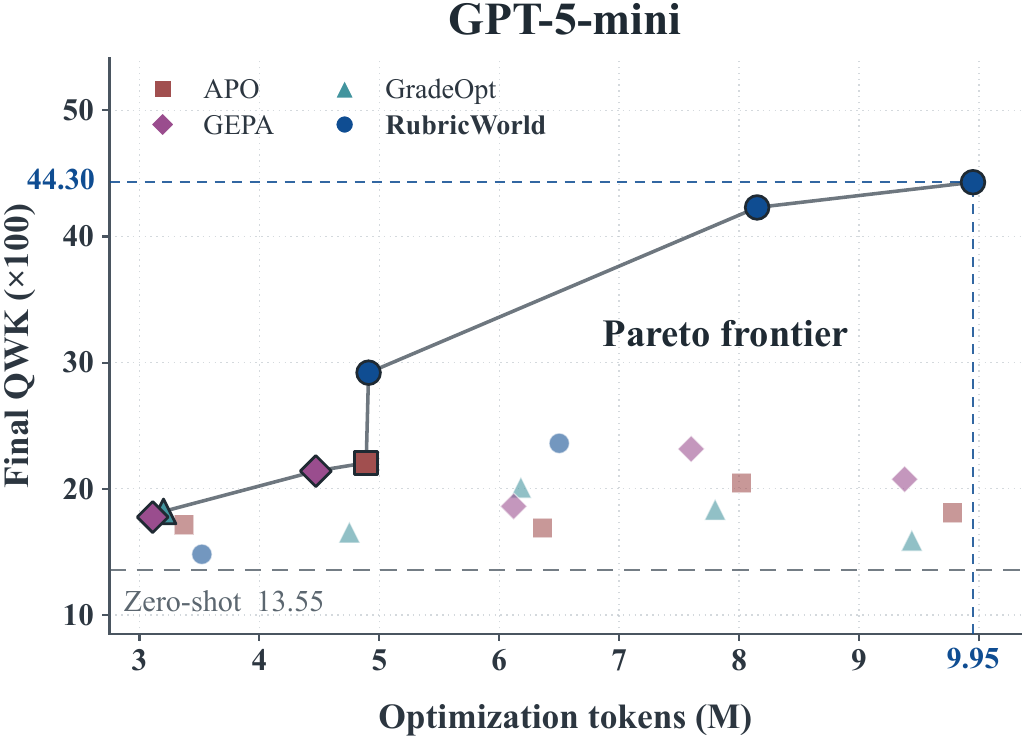}
\caption{\textbf{Quality--cost trade-off on ASAP2.0 (\texttt{gpt-5-mini}).}
Three-seed means including warm-start tokens; outlined points are Pareto-optimal.}
\label{fig:quality_cost}
\end{figure}

\begin{figure}[!htbp]
\centering
\includegraphics[width=0.91\columnwidth]{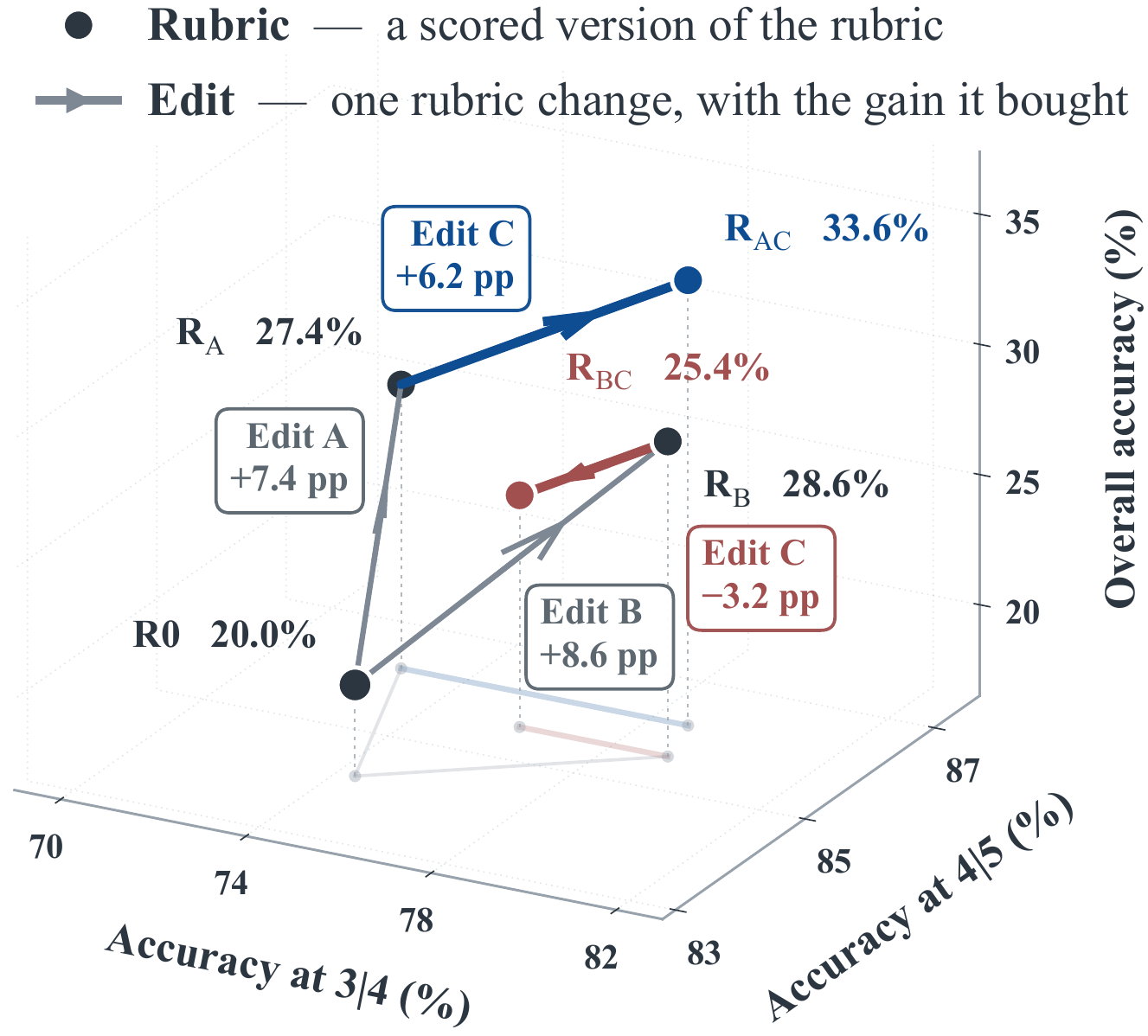}
\caption{\textbf{Illustrative rubric-editing trajectories (ASAP2.0 scale;
\texttt{gpt-5-mini} setting).}
The same edit C is applied after A or B.}
\label{fig:state_trajectory}
\end{figure}
\begin{figure*}[!t]
\centering
\begin{minipage}[t]{0.48\textwidth}
\centering
\includegraphics[width=\linewidth]{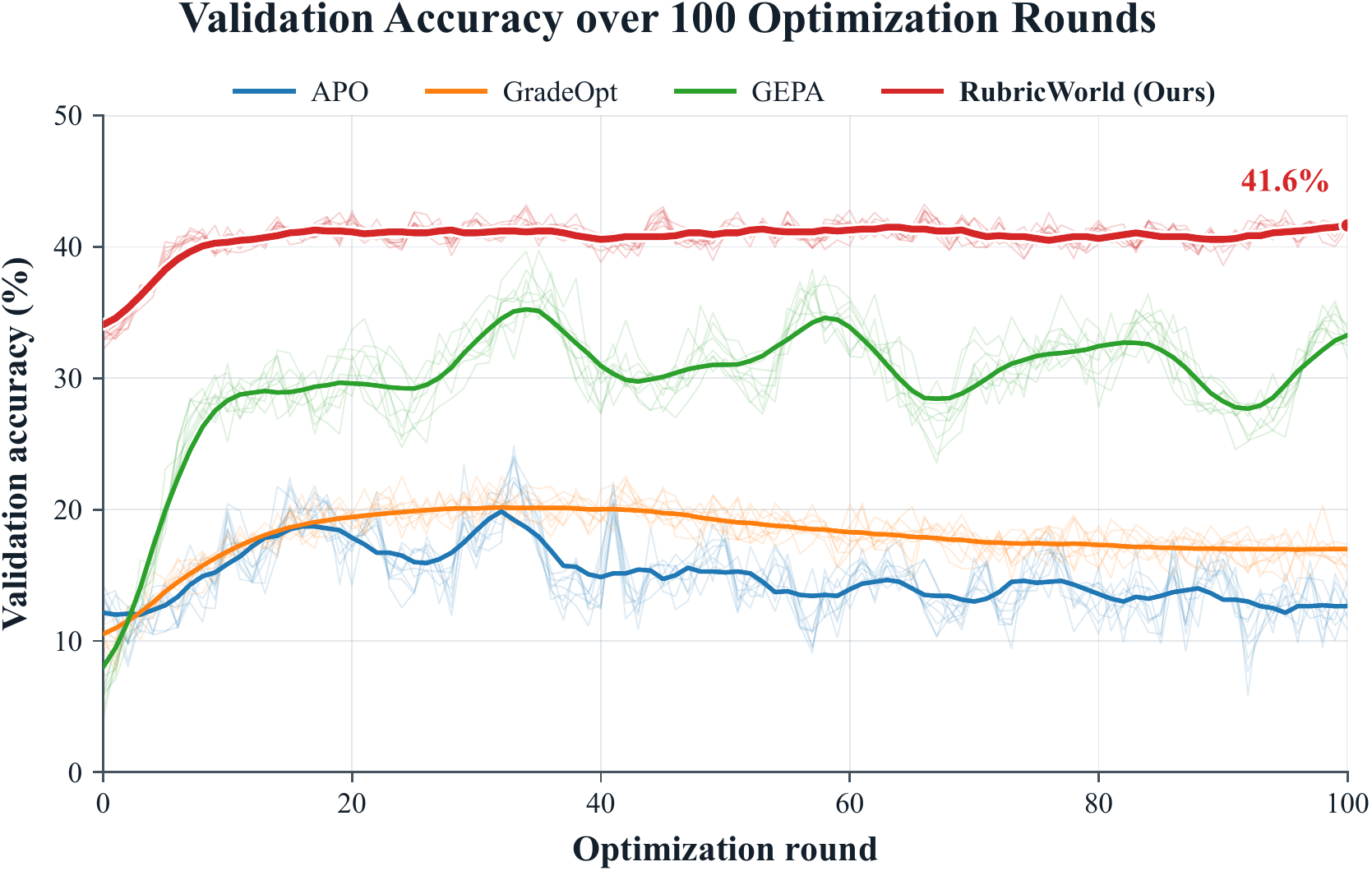}
\caption{\textbf{100-round validation ACC.}}
\label{fig:optimization_trajectory}
\end{minipage}\hfill
\begin{minipage}[t]{0.48\textwidth}
\centering
\includegraphics[width=\linewidth]{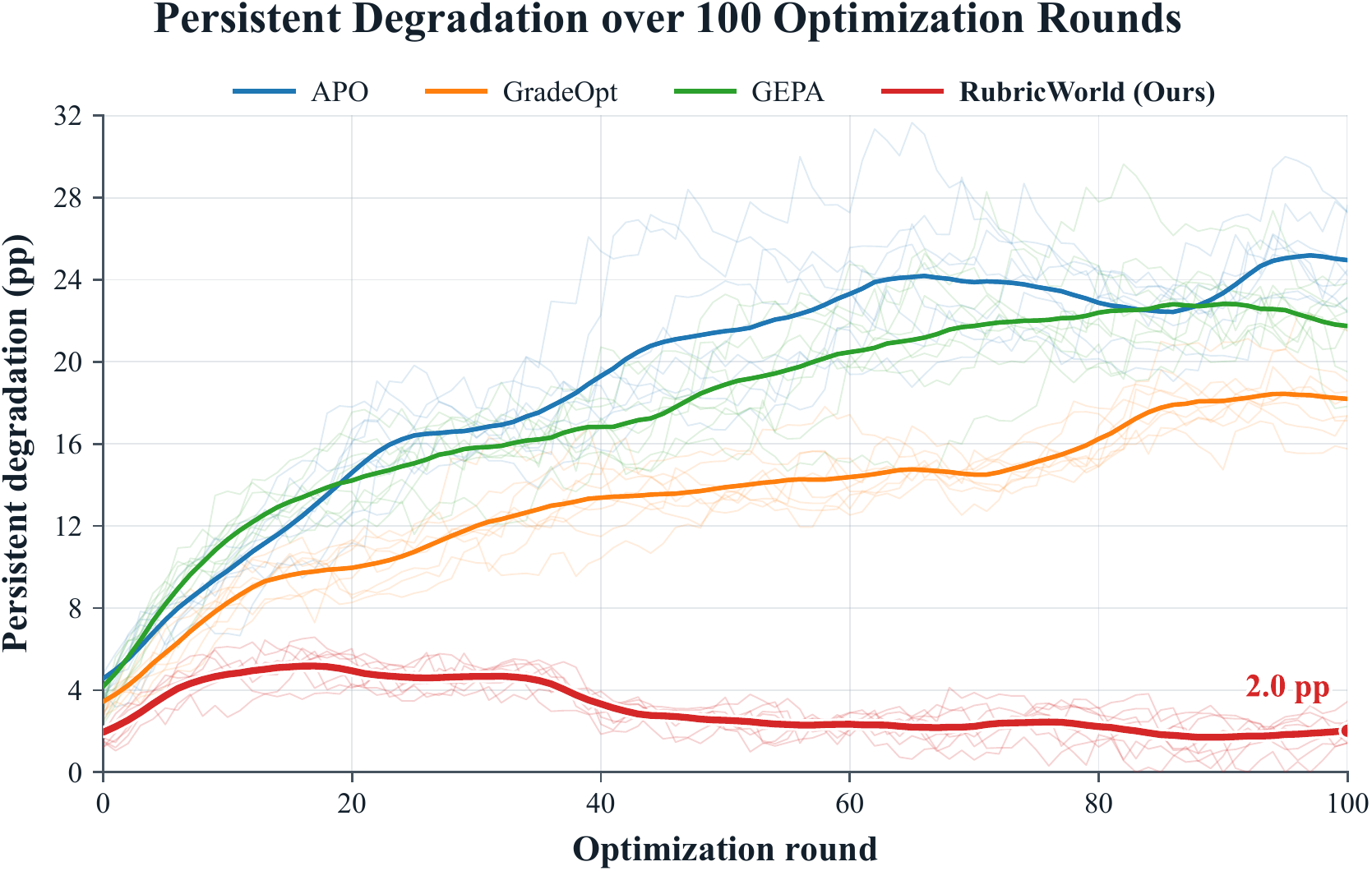}
\caption{\textbf{100-round persistent degradation (pp).}}
\label{fig:boundary_debt_trajectory}
\end{minipage}
\end{figure*}

\section{Results and Analysis}
\label{sec:results}

\subsection{Overall Grading Quality}
\label{sec:overall_results}

\method achieves the highest score in 20 of the 27 dataset--grader--metric
comparisons in Table~\ref{tab:main_results}.
Its strongest gains occur on ASAP2.0, where all nine comparisons favor
\method. QWK reaches 32.94, 37.25, and 45.10 with Llama-3.1-8B,
QwQ-32B, and \texttt{gpt-5-mini}, respectively; the strongest baselines
reach 8.92, 28.15, and 20.55.

The short-answer results are less uniform. On EIR with QwQ-32B,
APO has higher ACC (75.30\% versus 73.85\%) but lower QWK (71.85
versus 74.12), which penalizes larger scoring errors.
GradeOpt exceeds \method in ACC and QWK on EIR with
\texttt{gpt-5-mini}, and in ACC on ASAP-SAS with Llama-3.1-8B.
The gains therefore vary by dataset and metric.

\subsection{Grading Quality Versus Token Cost}
\label{sec:quality_cost}

Figure~\ref{fig:quality_cost} compares final QWK against optimization
tokens on ASAP2.0 with \texttt{gpt-5-mini}.
\method reaches 44.30 QWK near 10M tokens, with smaller differences
among methods at lower budgets. These token-constrained checkpoints
differ from the fixed five-round endpoint in
Table~\ref{tab:main_results} (45.10 QWK).

Fewer candidate evaluations need not mean lower total token cost
(Table~\ref{tab:world_model_e2e}). After twenty rounds, the full model
uses 8.32M tokens, compared with 8.68M without state and 8.15M with
direct evaluation. These totals include warm-start collection and do
not identify which calls account for the differences. The table's
quality measurements cover five rounds only.

\subsection{Why the Current Rubric Matters}
\label{sec:state_results}

Figure~\ref{fig:state_trajectory} illustrates how the same edit C can
have opposite effects after edits A and B.
Although the starting accuracies are close (27.4\% and 28.6\%),
C raises the first to 33.6\% and lowers the second to 25.4\%.
It corrects 41 answers and harms 10 on the A branch, compared with
7 corrections and 23 harms on the B branch.
At boundary $3|4$, correction/degradation rates are
29.9\%/2.8\% and 6.7\%/5.8\%, respectively.
This example motivates conditioning on the current rubric.
It does not establish how common the effect is or how well a model
predicts it; the axes show grading accuracy, not learned representations.
We assess predictive value separately on unseen rubric states.

\subsection{Prediction, Selection, and Acceptance}
\label{sec:ablation_results}

State conditioning improves prediction and candidate selection
(Figure~\ref{fig:ablation_triptych}a,b).
Compared with edit-only prediction, the full model lowers
$\mathrm{MAE}_C$ from 39.1 to 24.6\,pp and $\mathrm{MAE}_D$ from
34.5 to 19.8\,pp; additive history ties on $\mathrm{MAE}_D$.
The selected edits increase target correction from 8.5\% to 16.2\%
and reduce degradation from 8.8\% to 2.5\%.

The end-to-end ablation averages over the three graders
(Table~\ref{tab:world_model_e2e}). Five-round ACC/QWK is
40.18\%/38.43 for the full model, 35.42\%/33.16 without state,
and 27.31\%/24.85 with direct candidate evaluation and no learned
predictor (\LinkAppendix{app:prediction_variants}).
These matched-seed means do not establish statistical significance.

The acceptance rule trades some QWK for lower degradation
(Figure~\ref{fig:ablation_triptych}c).
QWK/degradation is 46.0/7.8\% for an empirical gate,
45.1/3.2\% for certification, and 41.2/14.2\% for prediction-only
acceptance. Thus certification reduces degradation by 4.6\,pp at a
cost of 0.9 QWK relative to the empirical gate.
These are test measurements, not certification bounds.

\subsection{Grading After 100 Refinement Rounds}
\label{sec:long_horizon_results}

Figures~\ref{fig:optimization_trajectory} and
\ref{fig:boundary_debt_trajectory} follow the same three ASAP2.0 runs
with \texttt{gpt-5-mini}.
\method stays near 41\% validation ACC while persistent degradation
falls toward two percentage points.
An earlier regression leaves this measure only when a later edit
corrects that boundary decision. Accuracy can therefore recover
while some earlier damage remains.
The two plots distinguish final grading quality from unrepaired
errors accumulated during editing; 

%% file: sections/table_main_results.tex
\newcommand{\std}[1]{\,{\tiny$\pm$#1}}

\begin{table*}[!t]
\centering

\scriptsize
\setlength{\tabcolsep}{2pt}
\renewcommand{\arraystretch}{1.14}
\begin{tabular*}{\textwidth}{@{\extracolsep{\fill}}l ccc ccc ccc@{}}
\toprule
\multirow{2}{*}{\textbf{Method}}
& \multicolumn{3}{c}{\textbf{EIR (National)}}
& \multicolumn{3}{c}{\textbf{ASAP2.0}}
& \multicolumn{3}{c}{\textbf{ASAP-SAS}} \\
\cmidrule(lr){2-4} \cmidrule(lr){5-7} \cmidrule(lr){8-10}
& \textbf{ACC} & \textbf{QWK} & \textbf{F1}
& \textbf{ACC} & \textbf{QWK} & \textbf{F1}
& \textbf{ACC} & \textbf{QWK} & \textbf{F1} \\
\midrule

\blockrow{Meta-Llama-3.1-8B-Instruct-AWQ-INT4} \\
Zero-shot \citep{brown2020language}
& 55.42\std{0.54} & 54.10\std{0.46} & 54.85\std{1.92}
& 12.50\std{2.15} & 5.75\std{2.13} & 4.55\std{0.70}
& 1.99\std{0.64} & 1.44\std{0.87} & 1.20\std{2.08} \\
Few-shot \citep{brown2020language}
& \gainone{57.12\std{2.00}} & \lossone{52.80\std{0.56}} & \gainone{55.15\std{0.20}}
& \lossone{9.40\std{0.54}} & \lossone{4.78\std{0.79}} & 4.74\std{1.45}
& \gaintwo{19.92\std{1.39}} & \lossone{0.01\std{2.25}} & \gainone{7.75\std{2.01}} \\
APO \citep{pryzant2023apo}
& \gaintwo{61.35\std{1.17}} & \gaintwo{60.15\std{1.97}} & \gaintwo{59.90\std{1.17}}
& \gainone{15.45\std{1.68}} & \gainone{8.92\std{1.27}} & \gainone{7.98\std{0.79}}
& \gaintwo{21.51\std{0.75}} & \gainone{8.90\std{1.56}} & \gaintwo{14.50\std{0.61}} \\
GradeOpt \citep{gradeopt}
& \lossone{54.20\std{0.83}} & \gainone{56.45\std{0.18}} & \lossone{53.10\std{1.99}}
& \lossone{10.48\std{0.68}} & \gainone{7.37\std{2.11}} & \gainone{5.44\std{1.53}}
& \gainthree{\textbf{33.07\std{1.76}}} & \gainone{9.15\std{1.53}} & \gaintwo{\textbf{21.09\std{0.43}}} \\
GEPA \citep{agrawal2025gepa}
& \gainone{58.90\std{2.09}} & \gainone{57.20\std{1.96}} & \gainone{57.45\std{0.31}}
& \gainone{13.10\std{1.31}} & \gainone{6.80\std{1.62}} & \gainone{5.90\std{0.67}}
& \gaintwo{16.40\std{0.96}} & \gainone{4.50\std{1.37}} & \gainone{8.20\std{1.62}} \\
\textbf{\method}
& \gainthree{\textbf{64.88\std{1.10}}} & \gainthree{\textbf{63.12\std{0.45}}} & \gainthree{\textbf{62.45\std{1.89}}}
& \gainthree{\textbf{34.94\std{0.52}}} & \gainthree{\textbf{32.94\std{1.90}}} & \gainthree{\textbf{30.77\std{0.30}}}
& \gainthree{29.50\std{1.65}} & \gainone{\textbf{9.40\std{1.22}}} & \gaintwo{20.15\std{0.36}} \\

\addlinespace[3pt]
\blockrow{Qwen/QwQ-32B-Preview-AWQ-INT4} \\
Zero-shot \citep{brown2020language}
& 68.15\std{1.46} & 67.40\std{2.20} & 67.90\std{0.91}
& 22.15\std{0.72} & 18.40\std{0.77} & 15.60\std{1.29}
& 14.20\std{1.28} & 11.05\std{0.43} & 10.80\std{1.03} \\
Few-shot \citep{brown2020language}
& \lossone{66.40\std{2.06}} & \gainone{68.95\std{2.10}} & \lossone{65.20\std{0.37}}
& \lossone{18.30\std{2.14}} & \lossone{12.60\std{0.62}} & \lossone{11.90\std{1.64}}
& \gaintwo{25.40\std{1.97}} & \losstwo{4.25\std{0.24}} & \gainone{15.60\std{1.47}} \\
APO \citep{pryzant2023apo}
& \gainthree{\textbf{75.30\std{1.10}}} & \gaintwo{71.85\std{0.93}} & \gainthree{\textbf{74.10\std{1.26}}}
& \gaintwo{31.40\std{2.07}} & \gaintwo{28.15\std{1.63}} & \gaintwo{25.90\std{2.13}}
& \gaintwo{26.80\std{1.65}} & \gainone{14.30\std{1.31}} & \gaintwo{21.15\std{2.02}} \\
GradeOpt \citep{gradeopt}
& \gainone{70.12\std{1.13}} & \gainone{69.50\std{2.05}} & \gainone{69.95\std{1.00}}
& \gainone{28.55\std{2.11}} & \gainone{22.10\std{0.34}} & \gainone{21.45\std{1.87}}
& \gainthree{33.45\std{1.06}} & \gaintwo{16.80\std{0.35}} & \gainthree{28.40\std{0.61}} \\
GEPA \citep{agrawal2025gepa}
& \gaintwo{72.05\std{1.54}} & \gainone{69.80\std{2.15}} & \gaintwo{71.15\std{2.20}}
& \gaintwo{26.80\std{1.24}} & \gainone{23.40\std{0.20}} & \gainone{20.25\std{2.31}}
& \gaintwo{22.10\std{1.92}} & \gainone{12.50\std{1.38}} & \gaintwo{17.30\std{0.68}} \\
\textbf{\method}
& \gaintwo{73.85\std{0.18}} & \gainthree{\textbf{74.12\std{1.96}}} & \gaintwo{72.60\std{1.97}}
& \gainthree{\textbf{38.90\std{1.35}}} & \gainthree{\textbf{37.25\std{1.39}}} & \gainthree{\textbf{35.10\std{0.76}}}
& \gainthree{\textbf{34.12\std{0.24}}} & \gainthree{\textbf{22.45\std{0.62}}} & \gainthree{\textbf{31.15\std{0.37}}} \\

\addlinespace[3pt]
\blockrow{OpenAI \texttt{gpt-5-mini}} \\
Zero-shot \citep{brown2020language}
& 73.42\std{1.87} & 72.85\std{0.50} & 73.11\std{0.78}
& 18.20\std{1.18} & 13.55\std{2.31} & 12.10\std{0.30}
& 7.80\std{1.97} & 5.15\std{0.97} & 5.45\std{1.72} \\
Few-shot \citep{brown2020language}
& \gainone{74.88\std{0.60}} & \gainone{74.12\std{1.88}} & \gainone{74.35\std{0.97}}
& \lossone{14.35\std{0.39}} & \lossone{10.20\std{0.34}} & \lossone{10.80\std{2.07}}
& \gaintwo{23.10\std{2.12}} & \lossone{2.45\std{1.06}} & \gainone{13.90\std{0.47}} \\
APO \citep{pryzant2023apo}
& \gaintwo{77.25\std{1.66}} & \gaintwo{76.90\std{2.05}} & \gaintwo{76.58\std{0.30}}
& \gainone{24.10\std{2.32}} & \gainone{20.55\std{2.23}} & \gainone{18.40\std{1.23}}
& \gaintwo{27.55\std{1.53}} & \gaintwo{14.20\std{2.31}} & \gaintwo{19.80\std{1.76}} \\
GradeOpt \citep{gradeopt}
& \gainthree{\textbf{82.94\std{0.49}}} & \gainthree{\textbf{83.12\std{2.15}}} & \gainthree{\textbf{82.85\std{0.27}}}
& \gainone{21.80\std{1.22}} & \gainone{15.40\std{1.85}} & \gainone{14.95\std{1.15}}
& \gainthree{35.20\std{1.80}} & \gaintwo{14.85\std{1.63}} & \gaintwo{25.60\std{1.84}} \\
GEPA \citep{agrawal2025gepa}
& \gaintwo{75.60\std{1.89}} & \gainone{74.90\std{0.85}} & \gaintwo{75.10\std{1.48}}
& \gainone{21.30\std{0.82}} & \gainone{17.65\std{1.28}} & \gainone{15.20\std{1.79}}
& \gaintwo{21.40\std{0.25}} & \gainone{10.30\std{0.35}} & \gaintwo{14.15\std{0.63}} \\
\textbf{\method}
& \gainthree{81.15\std{0.97}} & \gainthree{81.38\std{0.36}} & \gainthree{81.62\std{2.33}}
& \gainthree{\textbf{46.70\std{1.57}}} & \gainthree{\textbf{45.10\std{1.61}}} & \gainthree{\textbf{43.25\std{0.88}}}
& \gainthree{\textbf{39.15\std{0.85}}} & \gainthree{\textbf{19.85\std{1.63}}} & \gainthree{\textbf{29.70\std{2.27}}} \\
\bottomrule
\end{tabular*}
\caption{
\textbf{Final grading scores across three benchmarks and frozen LLM graders.}
Three-seed mean$\pm$SD; ACC/F1 in \%, QWK on a 0--100 scale.
Blue/red denotes changes from zero-shot; bold marks the best displayed score.
}
\label{tab:main_results}
\end{table*}

\begin{table}[t]
\centering

\scriptsize
\setlength{\tabcolsep}{3pt}
\renewcommand{\arraystretch}{1.12}
\begin{tabular}{lcccc}
\toprule
\multirow{2}{*}{\textbf{Variant}}
& \multirow{2}{*}{\textbf{QWK@5}$\uparrow$}
& \multirow{2}{*}{\textbf{ACC@5}$\uparrow$}
& \multicolumn{2}{c}{\textbf{Tokens (M)}$\downarrow$} \\
\cmidrule(lr){4-5}
& & & \textbf{5 rounds} & \textbf{20 rounds} \\
\midrule
\textbf{Full RubricWorld}
& 38.43 & 40.18 & 6.84 & 8.32 \\
w/o State
& 33.16 & 35.42 & 6.41 & 8.68 \\
w/o Learned Model
& 24.85 & 27.31 & 6.72 & 8.15 \\
\bottomrule
\end{tabular}
\caption{
\textbf{Edit-effect model ablation on ASAP2.0.}
Average across three frozen graders with shared generator and certification.
}
\label{tab:world_model_e2e}
\end{table}

%% file: sections/6_conclusion.tex
\Needspace{3\baselineskip}
\section{Conclusion}
\label{sec:conclusion}

\method learns corrections and degradations from paired grading
results to predict an edit's effect under the current rubric.
Predictions guide candidate selection; held-out evidence determines
acceptance. Rejected edits also contribute to learning.
Across three benchmarks and three frozen graders, the largest gains
occur on ASAP2.0. Ablations support the use of state information and
learned effects, while longer runs track errors left by earlier edits.

\section*{Limitations(does not count toward the main-text page limit)}
\label{sec:limitations}

\method relies on initial rubrics and human reference scores; high reference agreement does 
not inherently guarantee educational validity when human targets contain noise \citep{messick1995validity}. 
Empirical certification requires enough examples at checked boundaries;
rare score categories and small round-specific samples may prevent
otherwise useful edits from being certified.
The edit-effect model may not generalize to unfamiliar criteria or tasks.
Our evaluation covers educational grading, not open-domain judging,
safety evaluation, or temporal streaming responses
\citep{hashemi2024llmrubric}.

\section*{Ethical Considerations(does not count toward the main-text page limit)}
\label{app:ethics}

\method supports research on educational scoring; passing its audit does
not establish suitability for high-stakes grading. Incorrect scores can
affect students' opportunities, so consequential or disputed decisions
should receive human review, with a way to contest errors. Agreement with
reference scores does not establish fairness: references may contain bias,
and average metrics can hide unequal errors across groups. Use in practice
would require checking these differences in the intended population.
Student and teacher responses may also contain identifying information.
Sharing them, including with external model services, should respect access
conditions and minimize disclosure of personal information. These
obligations apply to both the private EIR (National) dataset and the public
ASAP2.0 and ASAP-SAS benchmarks (\LinkAppendix{app:data_access}).

%% file: sections/Appendix.tex
\section{AI Assistance Disclosure}
\label{app:ai_assistance}

ChatGPT assisted with language polishing (including the reduction of redundant and colloquial terms) and the layout design of figures and tables. The authors bear full responsibility for the scientific content, experimental results, references, and conclusions of the manuscript.

\section{Dataset Counts and Data Use}
\label{app:dataset_protocol}

Our essay tasks assign each response one ordered score, as in earlier automated essay scoring comparisons \citep{shermis2012contrasting}. Short-answer tasks use question-specific scoring criteria \citep{burrows2015eras}. The configuration specifies stratification by reference score when validation responses are selected from the training data. Unless otherwise stated, optimization uses only the training split and validation subsets; the final test set is held out until the final evaluation.

\subsection{EIR (National)}
\label{app:contents-eir}

EIR (National) contains five private short-answer items, each with an
item-specific score scale. To document the experimental split without
exposing restricted response content, Table~\ref{tab:eir_stats} reports
aggregate counts only.

\begin{table}[ht]
\centering
\small
\setlength{\tabcolsep}{3.5pt}
\renewcommand{\arraystretch}{1.06}
\begin{tabular}{lrccc}
\toprule
\textbf{Item} & \textbf{$n$} & \textbf{Scale} & \textbf{Train} & \textbf{Val/Test} \\
\midrule
U41Q1 & 199 & 0--3 & 159 & 20 / 20 \\
U41Q2 & 230 & 0--4 & 184 & 23 / 23 \\
U42Q1 & 248 & 0--4 & 198 & 25 / 25 \\
U42Q2 & 262 & 0--3 & 209 & 26 / 27 \\
U43Q1 & 434 & 0--4 & 347 & 43 / 44 \\
\midrule
Total & 1,373 & -- & 1,097 & 137 / 139 \\
\bottomrule
\end{tabular}
\caption{EIR (National) aggregate split statistics. Items use fixed 80/10/10
partitions, stratified by reference score where class counts permit.}
\label{tab:eir_stats}
\end{table}

No response appears in more than one partition, and the test partition is
reserved for final evaluation, consistent with the protocol in
\LinkSection{sec:experiments}.

\subsection{ASAP2.0}
\label{app:contents-asap2-0}

ASAP2.0 uses a 1--6 holistic scoring scale \citep{learningagency2024aes2}. Its essays come from the PERSUADE 2.0 corpus \citep{crossley2024persuade2}. The task requires evaluating student essays according to argument quality, evidence use, organization, language control, and mechanics. Table~\ref{tab:asap2_stats} reports the split statistics used in our experiments.

\begin{table}[ht]
\centering
\small
\setlength{\tabcolsep}{4pt}
\renewcommand{\arraystretch}{1.08}
\begin{tabular}{lrl}
\toprule
\textbf{Split} & \textbf{\#Samples} & \textbf{Score distribution} \\
\midrule
Train & 14,711 & 1058 / 4023 / 5326 / 3350 / 827 / 127 \\
Test  & 2,596  & 194 / 700 / 954 / 576 / 143 / 29 \\
\bottomrule
\end{tabular}
\caption{ASAP2.0 split statistics. Score distributions are ordered from score 1 to score 6.}
\label{tab:asap2_stats}
\end{table}

\paragraph{ASAP2.0 collection allocation.}
Recorded edit-effect collection uses split seed~0.
Of the 14,711 training answers, 1,500 are reserved for
$\mathcal D_{\mathrm{cert}}$ (10.2\%). We then sample 150 answers
for $\mathcal D_{\mathrm{dyn,val}}$ from the remaining 13,211,
followed by the dynamics-training sample from the remaining 13,061.
The 2,596 test answers remain unchanged and are used only for final
evaluation. Table~\ref{tab:asap2_collection} lists these allocations.

\begin{table}[ht]
\centering
\small
\setlength{\tabcolsep}{4pt}
\begin{tabular}{lr}
\toprule
Data role & Answers \\
\midrule
$\mathcal D_{\mathrm{dyn,train}}$ & 60 / 200 / 400 \\
$\mathcal D_{\mathrm{dyn,val}}$ & 150 \\
$\mathcal D_{\mathrm{cert}}$ & 1,500 \\
$\mathcal D_{\mathrm{test}}$ & 2,596 \\
\bottomrule
\end{tabular}
\caption{ASAP2.0 data allocation for edit-effect collection. The three
training counts are alternative collection budgets; the certification
count is the total pool size, not a per-round eligible sample count.}
\label{tab:asap2_collection}
\end{table}

The 60-answer sample is used for the initial diagnostic, the
200-answer sample for Pilot~A and prompt-order checks, and the
400-answer sample for the audit and main transition collection.
They are separate stratified samples from the same pool, not three
additional data partitions. Their identifiers are recorded separately.
The 150-answer validation sample is 1.1\% of its source pool; the
three training budgets are 0.5\%, 1.5\%, and 3.0\% of theirs.

\subsection{ASAP-SAS}
\label{app:contents-asap-sas}

ASAP-SAS consists of item-specific short-answer scoring tasks \citep{kaggle2013asapsas}. We use the public train/test split and carve out a validation set from the training data. The score scale is item-specific; for the reported Set 1 experiments, the score range is 0--3. We use Set 1 because its four score levels are sufficiently populated to support the three boundary-level diagnostics; we do not claim that this result transfers to the remaining items. Table~\ref{tab:asap_sas_stats} lists the training, validation, and test counts.

\begin{table}[ht]
\centering
\small
\setlength{\tabcolsep}{4pt}
\renewcommand{\arraystretch}{1.08}
\begin{tabular}{lrl}
\toprule
\textbf{Split} & \textbf{\#Samples} & \textbf{Score distribution} \\
\midrule
Train & 1,170 & 266 / 301 / 366 / 237 \\
Valid & 252   & Stratified from training data \\
Test  & 251   & 57 / 64 / 79 / 51 \\
\bottomrule
\end{tabular}
\caption{ASAP-SAS Set 1 split statistics. Score distributions are ordered from score 0 to score 3.}
\label{tab:asap_sas_stats}
\end{table}

\paragraph{Short-answer collection budgets.}
The dynamics-training and dynamics-validation budgets are 400 and
150 answers for ASAP-SAS, and 120 and 30 answers per EIR item.
These are internal collection subsets, distinct from the benchmark
train/validation/test counts above.

\subsection{Data Access: Private and Public Datasets}
\label{app:data_access}

EIR (National) is a private dataset and is not publicly available.
It contains written responses used to assess teachers' subject knowledge
and knowledge of teaching \citep{gradeopt}.
The other two datasets, ASAP2.0 and ASAP-SAS, are public benchmarks.
Their availability does not imply unrestricted reuse or redistribution;
users should follow the original providers' access and usage terms.

\section{Prompt and Patch Format}
\label{app:prompt_patch_format}

The scoring prompt is the text sent to the frozen grader. A patch proposes
an edit to that text and records why the change is needed. The formats below
show the initial essay rubric, the information supplied with a candidate edit,
and diagnostic paired-effect records; independent certification makes the
final commit decision (\LinkAppendix{app:certification_details}).

\subsection{Initial Scoring Prompt}
\label{app:contents-initial-scoring-prompt}

The initial prompt tells the grader what each essay score means. It is the
starting specification; optimization adds or revises scoring instructions
without changing the grader's parameters. This differs from approaches
that train rubric-following evaluators or construct explanatory evaluation
rubrics \citep{kim2024prometheus2,galvansosa2025rubriks}.

\begin{promptbox}{Initial Holistic Scoring Prompt}

You are grading a student essay response on a 1-6 holistic scale.

Rubric dimensions:
- Point of view and critical thinking
- Use of examples, reasons, and evidence
- Organization, focus, coherence, and progression
- Language quality, including vocabulary and sentence structure
- Grammar, usage, and mechanics

Score 6:
Clear and consistent mastery. The essay presents a strong position,
well-developed reasoning, relevant evidence, coherent organization,
and skillful language control.

Score 5:
Reasonably consistent mastery with occasional lapses.

Score 4:
Adequate mastery with lapses in quality.

Score 3:
Developing mastery. The essay shows partial control of the task but
has limited support, weak explanation, inconsistent organization,
or accumulating language problems.

Score 2:
Little mastery. The essay may contain a vague position or minimal
support, with weak organization and frequent language problems.

Score 1:
Very little or no mastery. The essay is highly limited, disorganized,
off-topic, or difficult to understand.

\end{promptbox}

\subsection{Boundary-Targeted Patch Schema}
\label{app:contents-boundary-targeted-patch-schema}

Each candidate contains the proposed scoring text and the evidence for it.
In the schema, \texttt{patch\_text} is the edit applied to the specification;
the other fields identify its target and explain the intended effect.

\begin{promptbox}{Boundary-Targeted Patch Schema}

{
  "target_boundary": "<score transition b|b+1>",
  "error_direction": "<over-scoring | under-scoring>",
  "criterion": "<affected rubric criterion>",
  "failure_cases": "<misgraded examples requiring correction>",
  "anchor_cases": "<correctly scored training examples>",
  "patch_text": "<proposed edit to the scoring instructions>",
  "expected_effect": "<intended boundary-level correction>",
  "rationale": "<why this update addresses the observed failure>"
}

\end{promptbox}

A patch for $3\mid4$ should clarify when a response deserves at least
score 4. Failure cases illustrate mistakes to repair, while training anchors are
correctly scored examples supplied to the generator. The intended scope guides candidate filtering, while paired-effect auditing
measures the resulting grading changes.

\paragraph{Illustrative edit.}
Suppose the selected group contains answers over-scored at $3\mid4$:
the grader gives at least 4, while their human scores are at most 3.
If these answers list evidence without explaining how it supports the
position, a candidate could add:
\begin{quote}
\small
To award a score of 4 or higher for evidence use, require an
explanation of how the examples or reasons support the stated
position. Listing examples alone does not meet this requirement.
\end{quote}
This example illustrates the format; it is not an edit recovered from
an experimental run. Its target is $3\mid4$, its direction is
over-scoring, and its criterion is evidence use. Correctly graded
training examples help the generator preserve answers that deserve
4. The edit may still lower some of those answers, so paired grading
measures both correction and degradation at every boundary before
acceptance is considered.

\subsection{Patch Generation Constraints}
\label{app:contents-patch-generation-constraints}

The generation instructions ask for an edit that addresses the observed
failure, retains the original criteria, and targets the selected score
distinction. These requests do not by themselves guarantee compliance. For example, an edit about explaining evidence
should not add a new requirement unrelated to the rubric.

We represent the candidate as $p=(b,d_{\mathrm{dir}},\mathcal F^{\mathrm{train}},\mathcal A^{\mathrm{train}},\Delta R)$:
$b$ is its intended boundary, $d_{\mathrm{dir}}$ is over- or under-scoring,
$\mathcal F^{\mathrm{train}}$ contains training failures,
$\mathcal A^{\mathrm{train}}$ contains training anchors, and $\Delta R$ is the
proposed text edit. The target metadata identifies intended corrections; paired grading
measures the effects of the proposed edit.

\subsection{Audit Metadata}
\label{app:contents-audit-metadata}

A selected candidate is associated with diagnostic paired-effect measurements
and a separate held-out certification outcome:

\begin{promptbox}{Patch Audit Record}

{
  "target_correction": "<paired target-boundary correction>",
  "boundary_degradation": "<paired degradation by boundary>",
  "repair_support": "<eligible paired counts by boundary>",
  "degradation_support": "<eligible paired counts by boundary>",
  "certification_repair_support": "<held-out target-error count>",
  "certification_degradation_support": "<held-out initially-correct counts by boundary>",
  "certification_status": "<certified | rejected | unsupported>",
  "commit_status": "<accepted | retained>"
}

\end{promptbox}

The paired edit-effect record contains correction and degradation rates
at the observed score boundaries. These measurements update the learned
predictor irrespective of the subsequent commit decision. The final
acceptance rule instead applies exact one-sided confidence bounds on the
unused certification split: a candidate is committed only when the target
correction lower bound and every degradation upper bound in the prespecified
certification scope meet their respective thresholds
(\LinkSection{sec:certification}; \LinkAppendix{app:full_certification_rule}).
Residual boundary regressions are tracked separately for long-horizon analysis
(\LinkAppendix{app:persistent_regressions}).

\newcolumntype{Y}{>{\centering\arraybackslash}X}

\tcbset{
  methodprompt/.style={
    enhanced,
    colback=black!2,
    colframe=black!35,
    boxrule=0.45pt,
    arc=1mm,
    left=1.5mm,
    right=1.5mm,
    top=1.2mm,
    bottom=1.2mm,
    before skip=0pt,
    after skip=0pt,
    fonttitle=\bfseries\small
  }
}

\section{Baseline Implementation Details}
\label{app:baseline_details}

The methods differ in what they add to or revise in the scoring
specification, and how they choose the resulting text. The grading LLM
is frozen for every method.

\begin{table}[t]
\centering
\small
\setlength{\tabcolsep}{3pt}
\renewcommand{\arraystretch}{1.12}
\begin{tabularx}{\columnwidth}{@{}l>{\raggedright\arraybackslash}X>{\raggedright\arraybackslash}X@{}}
\toprule
Method & Text supplied or changed & Selection basis\\
\midrule
Zero-shot & Original rubric & No update\\
Few-shot & Scored examples appended to prompt & Validation agreement\\
APO & Rewritten scoring instructions & Aggregate validation score\\
GradeOpt & Guidelines and examples derived from errors & Validation improvement\\
GEPA & Prompt variants generated from reflection & Candidate-pool selection\\
\method & Edit to a chosen score distinction & Repair and per-boundary breakage\\
\bottomrule
\end{tabularx}
\caption{What each comparison method changes and evaluates.}
\label{tab:baseline_details}
\end{table}

APO uses feedback on grading errors to propose revised instructions, then
selects by aggregate validation performance. GradeOpt turns misgraded
responses into revised guidelines and examples. GEPA uses reflection on
evaluation traces to generate prompt variants and maintains a pool of
candidates. \method evaluates an edit by the intended failures it repairs
and the previously correct decisions it breaks at each score boundary.
Their search strategies follow the original method descriptions, while the frozen grader is shared across comparisons.

\section{Hyperparameter Settings}
\label{app:hyperparameters}

Table~\ref{tab:hyperparameters} lists the reported settings. A round is one
cycle of target selection, candidate generation, and acceptance checking.
A patch count refers to proposed text edits, including rejected candidates.

\begin{table}[t]
\centering
\small
\setlength{\tabcolsep}{4pt}
\renewcommand{\arraystretch}{1.1}
\begin{tabular}{ll}
\toprule
Parameter & Value\\
\midrule
Optimization rounds & 5\\
Main-table repetitions & 3 seeds; mean$\pm$std\\
Long-horizon analysis rounds & 100\\
Candidate patches per round & 5\\
Validation sampling & By reference score\\
Data used to find errors & Training partition\\
Minimum target repair $\gamma$ & 0.80\\
Protected regression tolerance $\epsilon_b$ & 0.08\\
Intended patch scope & One score boundary\\
\bottomrule
\end{tabular}
\caption{Reported optimization settings for \method.}
\label{tab:hyperparameters}
\end{table}

\section{Audit Threshold Selection}
\label{app:audit_threshold}

The two thresholds constrain different one-sided confidence bounds.
$\gamma=0.80$ is the minimum required lower bound for target-boundary
correction; $\epsilon_b=0.08$ is the maximum allowed upper bound for
degradation on each boundary in the prespecified certification scope.
For illustration, a sample with 20 target failures and 100 initially
correct decisions would have empirical point estimates of 80\% correction
and 8\% degradation after 16 repairs and eight regressions. Certification,
however, evaluates the corresponding confidence limits rather than these
point estimates (\LinkAppendix{app:exact_binomial_bounds}).

The thresholds are fixed before final test evaluation and shared across
datasets and graders; improvements at one boundary do not offset a bound
violation at another certified boundary.

\section{Data Isolation and Candidate Construction}
\label{app:data_and_candidates}
\label{app:candidate_generation}

This section provides the data-isolation protocol and candidate
construction procedure used by the optimization loop in
Section~\ref{sec:method}.

\subsection{Data Isolation}
\label{app:data_isolation}

Each data point is a student or teacher response with its human score
for a fixed scoring task. We reserve the final test partition and assign
the remaining labeled answers to three separate optimization roles,
yielding four disjoint answer sets:
\[
\mathcal D_{\mathrm{dyn,train}},\quad
\mathcal D_{\mathrm{dyn,val}},\quad
\mathcal D_{\mathrm{cert}},\quad
\mathcal D_{\mathrm{test}}.
\]
Training responses supply candidate construction, paired-effect
measurements, and model fitting. The dynamics-validation set is
reserved for predictor tuning; the certification set governs commit
decisions and the final test set is accessed only for final evaluation.

\paragraph{Identifiers and sampling order.}
An answer is identified by a SHA-1 hash of its item and response text,
$\mathrm{sha1}(\mathrm{item}\Vert\mathrm{text})$, rather than its row
number. We reserve the supplied test split, then the certification
pool. Cross-pool duplicates are removed with priority given to test,
then certification, then dynamics data. From the remaining training
pool, we draw dynamics-validation answers first and dynamics-training
answers second, both stratified by human score, using seeds
\texttt{seed} and \texttt{seed+1}, respectively.

The split loader raises an error if an identifier remains in more
than one pool. Certification and test records are also rejected as
world-model training inputs. These checks detected cross-pool
duplicates in EIR; duplicate removal therefore precedes sampling.

The answer partitions are distinct from the world model's training
records: a model record contains a state, an edit, and aggregate rates
measured on a batch of training answers. Reference-score stratification
is used when validation answers are drawn from training data, as
described in \LinkAppendix{app:dataset_protocol}.

\textbf{Grouped prediction evaluation is a separate procedure.}
Within the dynamics-training data, we separately form held-out folds
by \texttt{edit\_id} (edit-instance groups), \texttt{edit\_text}
(unseen clauses), and \texttt{rubric\_state\_id} (unseen states).
Every example from a held-out group stays outside that fold's model
training data; group-level results are not used for certification.
This grouping tests prediction generalization and does not replace
the four dataset partitions or the final test holdout.

Partition membership is fixed before optimization.
Responses used for response-model fitting or candidate construction are
never reused as certification examples.
Likewise, no information from $\mathcal D_{\mathrm{test}}$ is used for
candidate generation, response prediction, candidate selection, or
certification.

When duplicate response texts occur in the original data, duplicates
are assigned to the same partition before optimization.
This prevents identical responses from entering both the model-training
and certification pathways.

\subsection{Round-Wise Certification Splits}
\label{app:roundwise_certification}

For a run containing $T$ optimization rounds, the certification data
are partitioned in advance into disjoint subsets:
\[
    \mathcal D_{\mathrm{cert}}
    =
    \bigcup_{t=0}^{T-1}
    \mathcal D_{\mathrm{cert}}^{(t)},
\]
with
\[
    \mathcal D_{\mathrm{cert}}^{(t)}
    \cap
    \mathcal D_{\mathrm{cert}}^{(t')}
    =
    \varnothing
    \qquad
    \text{for }t\neq t'.
\]

Round $t$ accesses only
$\mathcal D_{\mathrm{cert}}^{(t)}$.
Therefore, certification outcomes from earlier rounds cannot reveal
examples that will later be used to certify another update.

This distinction is important because candidate generation and response
prediction adapt to earlier optimization outcomes.
Fresh round-specific certification subsets keep the final commit
decision separate from this adaptive optimization process.
The outer benchmark split counts in \LinkAppendix{app:dataset_protocol}
are distinct from the eligible before-edit correction and degradation
counts used in each certificate. Those round-level denominators are
$n_t^C$ and $n_{t,b}^{D}$ in \LinkAppendix{app:certification_counts}.

\subsection{Certification Scope}
\label{app:certification_scope}

For contiguous score labels from $y_{\min}$ to $y_{\max}$, let
\[
    \mathcal B=\{y_{\min},\ldots,y_{\max}-1\}
\]
denote the actual score boundaries, including $0\mid1$ on a 0--3 item.
Finite-sample guarantees are reported only over a pre-specified subset
\[
    \mathcal B_{\mathrm{cert}}
    \subseteq
    \mathcal B.
\]

The certification scope $\mathcal B_{\mathrm{cert}}$ specifies which
boundaries the experiment seeks to protect with statistical evidence.
It is fixed before candidate outcomes are inspected; boundaries cannot
be removed after observing poor results or insufficient support.
Each required confidence bound is then checked using the corresponding
eligible certification decisions.

Boundaries outside $\mathcal B_{\mathrm{cert}}$ remain part of the edit
response and are still considered by the response model and candidate
selection rule.
They are reported empirically, but no finite-sample certification claim
is made for them.

\subsection{Error Groups}
\label{app:error_groups}

At optimization round $t$, the frozen grader evaluates
$\mathcal D_{\mathrm{dyn,train}}$ using the current scoring specification
$R_t$.
Each incorrect prediction is associated with the score boundaries that
separate the predicted score from its reference score.

Candidate construction operates on localized error groups.
We write a group as
\[
    g=(b,d_{\mathrm{dir}},o),
\]
where $b$ denotes the affected score boundary,
$d_{\mathrm{dir}}$ is the over- or under-scoring direction, and
$o$ records whether the error was already present or appeared after an
earlier accepted revision.

Over-scoring at $b$ has $\hat z_{i,b}^{(t)}=1$ and $z_{i,b}=0$;
under-scoring has $\hat z_{i,b}^{(t)}=0$ and $z_{i,b}=1$.
There are at most $2(K-1)$ boundary--direction groups, or
$4(K-1)$ after separating errors present under the initial rubric
from errors introduced by accepted edits. Empty groups are omitted.
An answer with an error spanning multiple boundaries contributes to
each affected group.

The origin field is used to describe the observed optimization history;
it is not treated as a causal explanation of the error.

Let $\mathcal E_t$ be the set of training-answer grading errors at
round $t$, and let $\mathcal G_t$ be the groups formed from those
errors. The routing rule selects
\[
    g_t=\rho_t(\mathcal E_t)
       =(b_t^\star,d_{t,\mathrm{dir}},o_t)
       \in\mathcal G_t
\]
for candidate construction.
The corresponding boundary $b_t^\star$ becomes the target boundary for
that optimization round.
\paragraph{Selection order.}
We first exclude groups whose associated rubric-clause family has
already failed twice. Groups smaller than the configured minimum
size are set aside unless no larger eligible group remains. We then
select the largest remaining group. Ties are resolved by increasing
boundary, followed by direction name and origin in lexical order.
This rule uses no random sampling, so identical error records give
the same selection. The origin field follows the two categories
defined above: errors present initially and errors introduced later.

\subsection{Scoped Candidate Generation}
\label{app:scoped_candidate_generation}

The candidate generator receives the current scoring specification
$R_t$, examples from the selected error group, and relevant correctly
graded responses from $\mathcal D_{\mathrm{dyn,train}}$.

The signed error direction first determines the intended function of
the revision.
Depending on the observed failure, the candidate may tighten a scoring
distinction, loosen it, or clarify an ambiguous decision rule.
The target and direction are determined before the final text of the
candidate clause is generated.

The generator then produces a small collection
\[
    \mathcal P_t
    =
    \{p_t^1,\ldots,p_t^M\}
\]
of localized revisions.
Each $p_t^j$ describes a textual modification to $R_t$ rather than an
entire replacement rubric.

Candidates are filtered before response prediction.
A candidate is removed if it merely duplicates an existing rule,
depends on the topic or wording of a specific response, introduces an
unrelated scoring requirement, directly contradicts the current
specification, or duplicates another candidate in the same pool.

Candidate generation itself never modifies the deployed scoring
specification.
For every candidate,
\[
    R_t\oplus p_t^j
\]
is a temporary specification until the candidate passes both
model-guided selection and empirical certification.

\section{Edit-Response Measurement Details}
\label{app:edit_response_details}

This section expands the paired edit-response definition introduced in
Section~\ref{sec:edit_effect}.

\subsection{Ordinal Boundary Representation}
\label{app:ordinal_boundaries}

For contiguous labels $y_{\min},\ldots,y_{\max}$,
$\mathcal B=\{y_{\min},\ldots,y_{\max}-1\}$ contains all $K-1$
thresholds. EIR items scored 0--3 use $0\mid1$, $1\mid2$, and $2\mid3$;
items scored 0--4 have four boundaries. Each score is represented by
$K-1$ binary decisions:
\[
    z_{i,b}
    =
    \mathbb I(y_i>b),
    \qquad
    b\in\mathcal B.
\]

Likewise, the prediction under the current rubric is
\[
    \hat z_{i,b}^{(t)}
    =
    \mathbb I(\hat y_i^{(t)}>b).
\]

After applying candidate edit $p$,
\[
    \hat z_{i,b}^{p}
    =
    \mathbb I(\hat y_i^{p}>b)
\]
denotes the corresponding decision under $R_t\oplus p$.

A single score change can modify several boundary decisions.
For example, moving from score 2 to score 4 changes both the
$2|3$ and $3|4$ decisions.
The boundary representation therefore preserves information that would
be lost if an edit were represented only by the scalar difference
between predicted scores.

\subsection{Repair and Regression Sets}
\label{app:repair_regression_sets}

Before applying candidate $p$, define
\[
    \mathcal F_{t,b}
    =
    \left\{
        i:
        z_{i,b}
        \neq
        \hat z_{i,b}^{(t)}
    \right\},
\]
and
\[
    \mathcal A_{t,b}
    =
    \left\{
        i:
        z_{i,b}
        =
        \hat z_{i,b}^{(t)}
    \right\}.
\]

$\mathcal F_{t,b}$ contains decisions that are currently wrong at
boundary $b$.
$\mathcal A_{t,b}$ contains decisions that are currently correct at the
same boundary.

For nonempty $\mathcal F_{t,b}$, target repair is measured as
\[
    C_{t,b}(p)
    =
    \frac{1}{|\mathcal F_{t,b}|}
    \sum_{i\in\mathcal F_{t,b}}
    \mathbb I
    \left(
        z_{i,b}
        =
        \hat z_{i,b}^{p}
    \right).
\]

For nonempty $\mathcal A_{t,b}$, regression is measured as
\[
    D_{t,b}(p)
    =
    \frac{1}{|\mathcal A_{t,b}|}
    \sum_{i\in\mathcal A_{t,b}}
    \mathbb I
    \left(
        z_{i,b}
        \neq
        \hat z_{i,b}^{p}
    \right).
\]

Thus, $C_{t,b}$ measures errors repaired by the edit, while
$D_{t,b}$ measures correct decisions damaged by the same edit.

\subsection{Validity Indicators}
\label{app:response_masks}

Some response components can be undefined because their denominator is
empty.
We therefore define
\[
    m_{t,b}^{C}
    =
    \mathbb I
    \left(
        |\mathcal F_{t,b}|>0
    \right),
\]
and
\[
    m_{t,b}^{D}
    =
    \mathbb I
    \left(
        |\mathcal A_{t,b}|>0
    \right).
\]

The complete validity mask is
\[
    \mathbf m_t
    =
    \left[
        \mathbf m_t^{C};
        \mathbf m_t^{D}
    \right].
\]

\textbf{Incomplete-record drop rule.}
We retain an answer--edit evaluation only if the reference,
before-edit, and after-edit decisions are available at all $K-1$
boundaries. If any boundary entry is missing, the entire record is
dropped; no missing entry is padded or filled with zero.
This differs from an empty correction/degradation denominator:
such a well-formed transition is retained, but the undefined
aggregate rate coordinate receives a zero mask and is excluded
from the relevant prediction loss.

\subsection{Response Vector}
\label{app:response_vector}

The response associated with candidate edit $p$ is
\[
    \mathbf e_t(p)
    =
    \left[
        \mathbf C_t(p);
        \mathbf D_t(p)
    \right],
\]
where
\[
    \mathbf C_t(p)=[C_{t,b}(p)]_{b\in\mathcal B},
\]
and
\[
    \mathbf D_t(p)=[D_{t,b}(p)]_{b\in\mathcal B}.
\]

The response dimension is therefore
\[
    d_e=2(K-1).
\]
The coordinates follow the score boundaries in $\mathcal B$ for the
scoring task under consideration. Thus $d_e=10$ for a 1--6 scale,
$d_e=8$ for a 0--4 scale, and $d_e=6$ for a 0--3 scale.

For a candidate targeting $b_t^\star$,
\[
    C_{t,b_t^\star}(p)
\]
is the intended repair coordinate.
The remaining repair and regression coordinates describe effects away
from the target boundary.

The boundary vector is therefore used as a coordinate system for the
effect of an edit rather than as a separate optimization objective.

\subsection{Transition Records}
\label{app:transition_records}

Whenever a selected candidate is evaluated on
$\mathcal D_{\mathrm{dyn,train}}$, we record
\[
    \left(
        s_t,
        p_t^\star,
        \mathbf e_t^{\mathrm{obs}},
        \mathbf m_t^{\mathrm{obs}}
    \right).
\]

This record is retained even when the candidate is later rejected by
the certification stage.
Certification controls whether the rubric changes; it does not alter
the fact that the paired edit response has been observed.

The response history therefore evolves as
\[
    \mathcal H_{t+1}
    =
    \mathcal H_t
    \cup
    \left\{
        \left(
            p_t^\star,
            \mathbf e_t^{\mathrm{obs}},
            \mathbf m_t^{\mathrm{obs}}
        \right)
    \right\}.
\]

The supervised transition set becomes
\[
    \mathcal T_{t+1}
    =
    \mathcal T_t
    \cup
    \left\{
        \left(
            s_t,
            p_t^\star,
            \mathbf e_t^{\mathrm{obs}},
            \mathbf m_t^{\mathrm{obs}}
        \right)
    \right\}.
\]

Candidates that are never applied to
$\mathcal D_{\mathrm{dyn,train}}$ have no observed transition and therefore
do not enter $\mathcal H_t$ or $\mathcal T_t$.

\subsection{Persistent Regressions}
\label{app:persistent_regressions}

For trajectory-level analysis, we additionally track regressions that
remain active after subsequent optimization rounds.

A decision becomes an active regression when an accepted edit changes a
previously correct boundary decision into an incorrect one.
It remains active until a later accepted edit restores that decision.

This diagnostic distinguishes transient damage from errors that persist
through the optimization trajectory.
It is used only for analysis and is not an additional term in the
round-level optimization objective.

\section{State-Conditioned Edit-Effect Model Details}
\label{app:response_model_details}
\label{app:state_action_representation}

This section specifies the edit-effect model, its response targets,
and the transition data used for supervised updating.

\subsection{State Definition}
\label{app:state_definition}

The optimization state at round $t$ is
\[
    s_t=(R_t,\mathcal H_t,\mathbf v_t),
\]
where $R_t$ is the active scoring rubric, $\mathcal H_t$ stores
previously measured edit effects, and $\mathbf v_t$ summarizes the
frozen grader's behavior on labeled training answers.
Only $R_t$ conditions the frozen grader; $\mathcal H_t$ and $\mathbf v_t$
provide predictive information to the edit-effect model.
The supervised transition set $\mathcal T_t$ is distinct from the
state history $\mathcal H_t$.

For each training answer $x_i$, let $\pi^{(t)}_{i,k}$ denote its
available score probability for label $k$. Define its cumulative
probability above boundary $b$, threshold margin, and correctness as
\[
\begin{aligned}
q_{i,b}^{(t)} &= \sum_{k>b}\pi^{(t)}_{i,k},\\
r_{i,b}^{(t)} &= \left|q_{i,b}^{(t)}-\tfrac12\right|,\quad
c_{i,b}^{(t)} =\mathbb I(\hat z_{i,b}^{(t)}=z_{i,b}).
\end{aligned}
\]
For a fixed scoring task and grader, we pool answer features and
$\mathbf c_i$ into the fixed-dimensional summary $\mathbf v_t$,
adding $\mathbf q_i$ and $\mathbf r_i$ only when
the grader provides usable score probabilities. In particular,
API-only graders need not expose per-label probabilities: missing
probability and margin features are omitted, not inferred from token
logits. All reference-based features use training labels only;
neither certification nor test labels enter the state encoder.

\subsection{Edit and State Encoders}
\label{app:encoders}

The candidate-edit encoder is a deterministic mapping
\[
    \phi:
    p
    \mapsto
    a_p
    \in
    \mathbb R^{d_a}.
\]

We use a general-purpose text encoder to represent the proposed
clause together with its target boundary and correction direction:
\[
    a_p=\phi(p)=\operatorname{Enc}(p,b^\star,d_{\mathrm{dir}}).
\]
The embedding describes the edit's content and intended target.
Its effect on grading is learned from the paired measurements, rather
than inferred from the embedding alone.
The text encoder is separate from the frozen grader and does not
require its parameters or internal activations.
The text encoder is frozen; only the world-model predictor is trained.
Feature extraction is deterministic. Representations are L2-normalized,
computed in 32-bit floating point with batches of 32, and cached on
disk. No encoder is fine-tuned on certification or test data.

The state encoder $\psi(s_t)$ combines a text embedding of the current
rubric with historical edit measurements and the training-answer
summary $\mathbf v_t$, giving $h_t\in\mathbb R^{d_h}$.
History provides evidence about earlier attempts; the answer summary
describes current grading behavior.
Score-probability and margin features are included only when the grader
provides them, as described in \LinkAppendix{app:state_definition}.
Thus API-only graders can use text embeddings, returned scores, and
training labels without exposing model weights or score probabilities.
The grader is never retrained.
The predictor and end-to-end ablations use the same edit embedding
so that changing state inputs or removing prediction does not also
change the edit representation.

\subsection{Factorized Edit--State Interaction}
\label{app:factorized_interaction}

For a general state--edit pair $(s,p)$, let
\[
    a_p=\phi(p),
    \qquad
    h_s=\psi(s).
\]

The response logits are
\[
    \boldsymbol\eta_\theta(s,p)
    =
    W_a a_p
    +
    U
    \left[
        (A a_p)
        \odot
        (B h_s)
    \right]
    +
    c.
\]

The predicted response is
\[
    \boldsymbol\mu_\theta(s,p)
    =
    \sigma
    \left(
        \boldsymbol\eta_\theta(s,p)
    \right),
\]
where $\sigma$ is applied element-wise.

The parameters have dimensions
\[
    W_a
    \in
    \mathbb R^{d_e\times d_a},
\]
\[
    A
    \in
    \mathbb R^{r_{\mathrm{int}}\times d_a},
    \qquad
    B
    \in
    \mathbb R^{r_{\mathrm{int}}\times d_h},
\]
\[
    U
    \in
    \mathbb R^{d_e\times r_{\mathrm{int}}},
    \qquad
    c
    \in
    \mathbb R^{d_e}.
\]

The term
\[
    W_a a_p
\]
captures the component of the response associated with the candidate
edit independently of the current state.

The interaction term
\[
    U
    \left[
        (A a_p)
        \odot
        (B h_s)
    \right]
\]
captures how the same edit representation changes under different
states.

The two projected vectors
\[
    A a_p,
    \qquad
    B h_s
    \in
    \mathbb R^{r_{\mathrm{int}}}
\]
are combined through element-wise multiplication.
The resulting interaction features are mapped by $U$ into the
repair--regression response space.

A full three-way interaction over edit, state, and response dimensions
would require parameters proportional to
$d_a d_h d_e$.
The factorized form instead uses a shared interaction dimension $r_{\mathrm{int}}$ and
avoids explicitly constructing that tensor.

\subsection{Edit-Only Model}
\label{app:edit_only_model}

To test whether the current rubric state contributes predictive
information beyond the candidate edit itself, we remove the
edit--state interaction and use
\[
    \boldsymbol\mu_{\mathrm{edit}}(p)
    =
    \sigma
    \left(
        W_a a_p+c
    \right).
\]

The response targets, edit representation, transition data, and
training procedure remain unchanged.
The difference between this model and the full model therefore isolates
the value of conditioning on the current optimization state.

\subsection{Transition Training Set}
\label{app:transition_training}

Before round $t$, the response model has access to
\[
    \mathcal T_t
    =
    \left\{
        (
            s_n,
            p_n,
            \mathbf e_n,
            \mathbf m_n
        )
    \right\}_{n=1}^{N_t}.
\]

Each transition records the state from which an edit was evaluated, the
candidate edit, its measured paired response, and the validity mask for
that response.

An evaluated candidate remains in $\mathcal T_t$ regardless of whether
it was subsequently committed.
This allows the model to learn from unsuccessful or unsafe edits without
allowing those edits to modify the deployed scoring specification.

\subsection{Masked Response Loss}
\label{app:masked_loss}

Equation~\eqref{eq:effect_loss} gives the training objective.
The predicted and observed rate at coordinate $k$ are
$\mu_{\theta,k}(s,p)$ and $e_k$, respectively. The inner sum includes
only coordinates with $m_k=1$; the outer sum covers the measured
state--edit records in $\mathcal T_t$.

An empty correction or degradation denominator leaves its rate
undefined. The mask removes that coordinate instead of treating it
as zero. The sigmoid constrains all predicted rates to $[0,1]$, the
same range as the observed rates.

\subsection{Warm-Start Transitions}
\label{app:warm_start}

The model is initialized from paired edit-effect measurements on
$\mathcal D_{\mathrm{dyn,train}}$, collected before model-guided
selection. The number of answer--boundary observations is
\[
 N_{\mathrm{boundary}}=
 \sum_{(s,p)\in\mathcal T_0}n_{s,p}(K_{s,p}-1),
\]
where $n_{s,p}$ counts complete answer--edit records for a measured
state--edit intervention. Under a balanced design, this becomes
$N_{\mathrm{state}}N_{\mathrm{edits/state}}N_{\mathrm{answers/edit}}(K-1)$.
Our six-state pilot collected 36,000 boundary records, corresponding
to 7,200 complete answer--edit records for the five-boundary scale.
Boundary records are not treated as independent rubric trajectories.
The additional observations from selected edits, including rejected
edits, use the same training schema; $\mathcal D_{\mathrm{cert}}$ and
$\mathcal D_{\mathrm{test}}$ are never used to fit the model.

\textbf{State-held-out evaluation.}
Using \texttt{rubric\_state\_id}-grouped folds, the full state-conditioned model reaches $R^2=0.283$ on unseen
states, compared with $0.241$ for the edit-only model (a difference of
$0.042$).
The separate \texttt{edit\_id} and \texttt{edit\_text} groupings check
edit-instance and unseen-clause generalization; folds do not replace
the four dataset partitions defined above.

\subsection{Predictive Uncertainty}
\label{app:predictive_uncertainty}

We train $E=5$ independently seeded edit-effect models, following
deep-ensemble uncertainty estimation \citep{lakshminarayanan2017deepensembles}.
For candidate $p_t^j$, let $\boldsymbol\mu_e(s_t,p_t^j)$ denote the
prediction of ensemble member $e$. The ensemble mean and
per-coordinate sample standard deviation are
\[
\begin{aligned}
\bar{\boldsymbol\mu}
&=\frac{1}{E}\sum_{e=1}^{E}\boldsymbol\mu_e,\\
\boldsymbol\sigma
&=\sqrt{\frac{1}{E-1}\sum_{e=1}^{E}
(\boldsymbol\mu_e-\bar{\boldsymbol\mu})^2}.
\end{aligned}
\]
The selection rule uses $\bar{\boldsymbol\mu}$ as its predicted mean;
the standard deviation measures disagreement among models, not a
calibrated confidence interval.

The corresponding target-repair quantities are
\[
    \mu^C_{t,j,b_t^\star},
    \qquad
    \sigma^C_{t,j,b_t^\star},
\]
while the regression quantities are
\[
    \mu^D_{t,j,b},
    \qquad
    \sigma^D_{t,j,b}.
\]

These quantities are converted into
\[
    \widetilde C_{t,j}
    =
    \mu^C_{t,j,b_t^\star}
    -
    \beta
    \sigma^C_{t,j,b_t^\star},
\]
and
\[
    \widetilde D_{t,j,b}
    =
    \mu^D_{t,j,b}
    +
    \beta
    \sigma^D_{t,j,b}.
\]

The predictive mean and standard deviation affect candidate
ranking and filtering only. Even if the ensemble is miscalibrated,
it cannot bypass the separate certification gate:
commits rely on fresh $\mathcal D_{\mathrm{cert}}$ observations and
one-sided finite-sample bounds, subject to their sampling assumptions.
Miscalibration can impair which candidates are explored; it does not
itself provide authority to commit an edit.

\subsection{Online Model Updating}
\label{app:online_updating}

After selecting $p_t^\star$, the candidate is evaluated on
$\mathcal D_{\mathrm{dyn,train}}$ to obtain
\[
    \mathbf e_t^{\mathrm{obs}},
    \qquad
    \mathbf m_t^{\mathrm{obs}}.
\]

The response history becomes
\[
    \mathcal H_{t+1}
    =
    \mathcal H_t
    \cup
    \left\{
        (
            p_t^\star,
            \mathbf e_t^{\mathrm{obs}},
            \mathbf m_t^{\mathrm{obs}}
        )
    \right\},
\]
and the supervised transition set becomes
\[
    \mathcal T_{t+1}
    =
    \mathcal T_t
    \cup
    \left\{
        (
            s_t,
            p_t^\star,
            \mathbf e_t^{\mathrm{obs}},
            \mathbf m_t^{\mathrm{obs}}
        )
    \right\}.
\]

The response model is updated from
$\mathcal T_{t+1}$
before candidate selection at round $t+1$.

If no candidate satisfies the predicted constraints at round $t$, no
new paired transition is produced and
\[
    \mathcal H_{t+1}=\mathcal H_t,
    \qquad
    \mathcal T_{t+1}=\mathcal T_t.
\]

\section{Candidate Selection Details}
\label{app:selection_details}

This section expands the model-guided selection rule in
Section~\ref{sec:selection}.

\subsection{Conservative Predicted Response}
\label{app:conservative_response}

For candidate $p_t^j$ targeting boundary $b_t^\star$, define
\[
    \widetilde C_{t,j}
    =
    \mu^C_{t,j,b_t^\star}
    -
    \beta
    \sigma^C_{t,j,b_t^\star},
\]
and
\[
    \widetilde D_{t,j,b}
    =
    \mu^D_{t,j,b}
    +
    \beta
    \sigma^D_{t,j,b}.
\]

The candidate chosen by the model is
\[
    p_t^\star
    =
    \arg\max_{p_t^j\in\mathcal P_t}
    \widetilde C_{t,j}
\]
subject to
\[
    \widetilde D_{t,j,b}
    \leq
    \epsilon_b,
    \qquad
    \forall b\in\mathcal B.
\]

The subtraction on repair penalizes uncertain predicted benefit.
The addition on regression penalizes uncertain predicted harm.

These quantities are used only to determine which candidate should be
measured with the frozen grader.
They do not determine whether that candidate is committed.

\subsection{No-Feasible-Candidate Case}
\label{app:no_feasible_candidate}

If
\[
    \left\{
        p_t^j\in\mathcal P_t:
        \widetilde D_{t,j,b}
        \leq
        \epsilon_b
        \ \forall b\in\mathcal B
    \right\}
    =
    \varnothing,
\]
the round produces no selected candidate.

In this case,
\[
    R_{t+1}=R_t,
\]
and no new transition is added to the response history.

The optimizer therefore does not relax its predicted regression
constraints merely to force an update at every round.

\section{Finite-Sample Certification}
\label{app:certification_details}

This section describes the statistical procedure used after a candidate
has been selected and measured.

\subsection{Certification Counts}
\label{app:certification_counts}

At round $t$, candidate $p_t^\star$ is evaluated on the previously
unused set
\[
    \mathcal D_{\mathrm{cert}}^{(t)}.
\]

For the target boundary $b_t^\star$, let
\[
    n_t^C
\]
be the number of certification examples that are incorrect before the
candidate edit and therefore eligible for repair.

Let
\[
    x_t^C
\]
be the number of those examples that become correct after the edit.

For each protected boundary $b$, let
\[
    n_{t,b}^{D}
\]
be the number of examples that are correct before the edit, and let
\[
    x_{t,b}^{D}
\]
be the number that become incorrect after the edit.

These denominators count the current rubric's decisions on the
round-specific certification subset. They are neither the total
certification-pool size nor the smaller human-label count on either
side of a boundary. The numerator requires the paired after-edit
score for the same answer. Thus pool sizes and label frequencies
alone do not establish a certificate.

For a fixed candidate and boundary, each eligible answer produces a
binary outcome: corrected or not, or degraded or not.
The exact binomial bounds assume independent answers with a common
event probability within each eligible population.
These assumptions are conditional on the current rubric, selected
candidate, and distribution used to sample certification answers.
The fresh round-specific set prevents this candidate from being chosen
using its own certification outcomes. It does not by itself establish
the sampling assumptions.
Dependence across boundaries is allowed by the subsequent union bound;
multiple boundary decisions from one answer are not pooled as
independent trials for a single rate.

\subsection{Exact One-Sided Binomial Bounds}
\label{app:exact_binomial_bounds}

For an observed count $x$ among $n$ Bernoulli trials, we use exact
one-sided binomial confidence bounds. They invert a binomial tail test;
the beta quantile below is a way to compute that bound, not a prior
distribution over edit effects. Here $\mathrm{Beta}^{-1}(q;a,b)$ denotes
the $q$-quantile of a beta distribution with shape parameters $a,b$.

For target repair, the lower Clopper--Pearson bound is
\begin{equation}
\label{eq:binomial_bounds}
    L^C(x,n;\alpha_C)
    =
    \mathrm{Beta}^{-1}
    \left(
        \alpha_C;
        x,
        n-x+1
    \right)
\end{equation}
for $x>0$, with
\[
    L^C(0,n;\alpha_C)=0.
\]

At round $t$,
\[
    L_{t,b_t^\star}^{C}
    =
    L^C
    \left(
        x_t^C,
        n_t^C;
        \alpha_C
    \right).
\]

For regression, the upper Clopper--Pearson bound is
\[
    U^D(x,n;\alpha_D)
    =
    \mathrm{Beta}^{-1}
    \left(
        1-\alpha_D;
        x+1,
        n-x
    \right)
\]
for $x<n$, with
\[
    U^D(n,n;\alpha_D)=1.
\]

Hence,
\[
    U_{t,b}^{D}
    =
    U^D
    \left(
        x_{t,b}^{D},
        n_{t,b}^{D};
        \alpha_{D,b}
    \right).
\]

\subsection{Certification Rule}
\label{app:full_certification_rule}

A candidate can be certified only when
\[
    b_t^\star
    \in
    \mathcal B_{\mathrm{cert}}.
\]

The repair requirement is
\[
    L_{t,b_t^\star}^{C}
    \geq
    \gamma.
\]

For every certified boundary,
\[
    U_{t,b}^{D}
    \leq
    \epsilon_b,
    \qquad
    \forall
    b\in
    \mathcal B_{\mathrm{cert}}.
\]

A candidate is committed only when all required inequalities hold.

If the rule succeeds,
\[
    R_{t+1}
    =
    R_t
    \oplus
    p_t^\star.
\]

Otherwise,
\[
    R_{t+1}
    =
    R_t.
\]

\subsection{Confidence Allocation Across Multiple Bounds}
\label{app:multiple_testing}

At each round, the certificate contains
$Q_{\mathrm{cert}}=1+|\mathcal B_{\mathrm{cert}}|$ one-sided statements:
one for target correction and one for degradation at each certified
boundary. Let $\delta$ be the family-wise error probability.

We allocate individual error probabilities
\[
    \alpha_1,\ldots,\alpha_{Q_{\mathrm{cert}}}
\]
such that
\[
    \sum_{q=1}^{Q_{\mathrm{cert}}}
    \alpha_q
    \leq
    \delta.
\]

We use a Bonferroni allocation to control simultaneous coverage, as
in classical family-wise error control \citep{holm1979multiple}.
A simple Bonferroni allocation sets
\[
    \alpha_q
    =
    \frac{\delta}{Q_{\mathrm{cert}}}.
\]

Under this construction, the probability that at least one reported
bound fails to cover its corresponding population quantity is at most
$\delta$ under the assumed sampling model.

A per-edit certificate allocates confidence only over the statements
included in that edit's certificate.
If a joint claim is made across multiple optimization rounds, the
confidence allocation must additionally account for the statements
made across those rounds.

\subsection{Minimum Sample Support}
\label{app:minimum_support}

A required quantity is classified as \emph{unsupported} when its sample
size is insufficient to establish the prescribed bound, even under the
most favorable possible observation.

For regression, consider the favorable case in which zero harmful
transitions are observed:
\[
    x=0.
\]

The exact upper bound reduces to
\[
    U^D(0,n;\alpha)
    =
    1-\alpha^{1/n}.
\]

To certify
\[
    D\leq\epsilon
\]
with zero observed regressions, $n$ must satisfy
\[
    1-\alpha^{1/n}
    \leq
    \epsilon.
\]

Equivalently,
\[
    n
    \geq
    \frac{\log\alpha}
         {\log(1-\epsilon)}.
\]

We therefore define
\[
    n_{\min}^{D}
    =
    \left\lceil
        \frac{\log\alpha}
             {\log(1-\epsilon)}
    \right\rceil.
\]

A regression coordinate with
\[
    n_{t,b}^{D}
    <
    n_{\min}^{D}
\]
cannot receive the requested certificate and is marked unsupported.

A corresponding minimum-support condition is applied to target repair:
the available number of target errors must be large enough that even a
perfect observed repair rate can produce
\[
    L^C
    \geq
    \gamma.
\]

\paragraph{Illustrative sample-size calculation.}
Suppose an edit is certified at three boundaries with $\delta=0.05$.
There are four statements: one target-correction bound and three
degradation bounds. Equal allocation gives $\alpha=0.05/4=0.0125$.
With $\epsilon=0.08$, zero observed degradations require
\[
    n_{\min}^{D}
    =\left\lceil\frac{\log(0.0125)}{\log(0.92)}\right\rceil
    =53
\]
initially correct decisions per checked boundary.
For correction, the best case is $x=n$, giving $L^C(n,n;\alpha)=\alpha^{1/n}$.
With $\gamma=0.80$, this requires
\[
    n_{\min}^{C}
    =\left\lceil\frac{\log(0.0125)}{\log(0.80)}\right\rceil
    =20
\]
initially wrong decisions at the target boundary, all corrected.
The two eligible sets at the target are disjoint, so this illustration
requires at least 73 certification answers in that round, even in the
best case. The actual requirement can be larger when corrections are
incomplete or degradations occur.

A joint claim over the same four statements in each of five rounds
would instead allocate $\alpha=0.05/(4\times5)=0.0025$, giving
$n_{\min}^{D}=72$ and $n_{\min}^{C}=27$ per round in the best case.
These are illustrations, not reported per-round sample sizes.
They show why an observed rate alone does not establish a certificate
and why the eligible counts must be reported for each experiment.

\subsection{Unsupported Outcomes}
\label{app:unsupported}

Certification produces three conceptually distinct outcomes.

\paragraph{Certified.}
All required repair and regression bounds are supported and satisfy the
specified thresholds.

\paragraph{Rejected.}
The required quantities have sufficient support, but at least one
confidence bound violates its threshold.

\paragraph{Unsupported.}
The target boundary lies outside the prespecified certification scope,
or at least one required quantity lacks sufficient sample support for
the prescribed bound.

Unsupported outcomes are never counted as certified.
They are reported separately from statistical rejection because the two
outcomes represent different reasons for withholding an update.

\subsection{Per-Edit and Trajectory-Level Guarantees}
\label{app:trajectory_guarantee}

A per-edit certificate applies only to the selected candidate and the
fresh certification sample used at that round.
It does not automatically provide a simultaneous guarantee for the
entire optimization trajectory.

A joint multi-round guarantee requires confidence allocation over all
statements included in that claim.
Our round-wise certification splits permit such a construction because
the certification observations for different rounds are separated in
advance.

When an experiment does not use the required fresh certification data
or joint confidence allocation, long-horizon results are described as
empirical trajectory analyses rather than finite-sample certified
guarantees.

\subsection{Scope and Interpretation}
\label{app:certification_interpretation}

The certification procedure addresses one narrow question:
whether the observed repair and regression rates provide sufficient
finite-sample evidence for the prescribed thresholds.

The certificate is conditional on the distribution from which the
certification examples are sampled and on the reference labels used to
define correctness.

It does not by itself establish educational validity, fairness,
distribution-shift robustness, or the absence of all possible future
grading errors.

The response model and certification procedure therefore play different
roles.
The former predicts which candidate is promising enough to evaluate;
the latter determines whether the measured candidate has sufficient
held-out evidence to be committed.

\section{Complete Optimization-State Update}
\label{app:complete_update}

For completeness, this section states the state update performed after
each optimization round.

If no candidate satisfies the predicted selection constraints,
\[
    R_{t+1}=R_t,
\]
\[
    \mathcal H_{t+1}
    =
    \mathcal H_t,
\]
and
\[
    \mathcal T_{t+1}
    =
    \mathcal T_t.
\]

If a candidate $p_t^\star$ is selected, it is first evaluated on
$\mathcal D_{\mathrm{dyn,train}}$.
Its observed transition is added to the history and supervised
transition data regardless of the later certification outcome:
\[
    \mathcal H_{t+1}
    =
    \mathcal H_t
    \cup
    \left\{
        (
            p_t^\star,
            \mathbf e_t^{\mathrm{obs}},
            \mathbf m_t^{\mathrm{obs}}
        )
    \right\},
\]
\[
    \mathcal T_{t+1}
    =
    \mathcal T_t
    \cup
    \left\{
        (
            s_t,
            p_t^\star,
            \mathbf e_t^{\mathrm{obs}},
            \mathbf m_t^{\mathrm{obs}}
        )
    \right\}.
\]

The certification outcome affects only the deployed scoring
specification:
\[
    R_{t+1}
    =
    \begin{cases}
        R_t\oplus p_t^\star,
        &
        \text{if }p_t^\star\text{ is certified},
        \\[4pt]
        R_t,
        &
        \text{otherwise}.
    \end{cases}
\]

Thus, an edit may contribute information to the learned response model
without being incorporated into the grader itself.
This separation allows rejected transitions to improve future
prediction while preventing uncertified edits from modifying the active
rubric.

\section{Evaluation Metric Details}
\label{app:metric_details}

\subsection{Edit-Response Prediction}

Let $\mathcal V$ denote the held-out collection of measured
state--edit transitions used for response-model evaluation.
For transition $(t,p)$ and boundary $b$, let
$m_{t,b}^{C}$ and $m_{t,b}^{D}$ denote the validity indicators defined
in Appendix~\ref{app:response_masks}.

Let $q=(t,p,b)$ index the held-out state--edit--boundary set
$\mathcal I=\{(t,p,b):(t,p)\in\mathcal V, b\in\mathcal B\}$.
Write $C_q=C_{t,b}(p)$, $D_q=D_{t,b}(p)$, and use the same
shorthand for predictions and validity masks. Then
\[
\mathrm{MAE}_{C}\;[\mathrm{pp}]=
100\,\frac{\sum_{q\in\mathcal I}m_q^C\lvert\widehat C_q-C_q\rvert}
{\sum_{q\in\mathcal I}m_q^C},
\]
\[
\mathrm{MAE}_{D}\;[\mathrm{pp}]=
100\,\frac{\sum_{q\in\mathcal I}m_q^D\lvert\widehat D_q-D_q\rvert}
{\sum_{q\in\mathcal I}m_q^D}.
\]

The factor of $100$ reports absolute errors in percentage points, as in
\LinkSection{sec:baseline_setup}; the underlying correction and degradation
rates themselves range from 0 to 1.
Undefined response coordinates are removed through the corresponding
validity indicators rather than assigned an artificial value.

\subsection{Candidate-Selection Metrics}

Let $\mathcal S$ denote the optimization rounds in which a candidate
edit is selected and evaluated.

The realized target repair is
\[
    \mathrm{Repair}_{\mathrm{sel}}
    =
    \frac{1}{|\mathcal S|}
    \sum_{t\in\mathcal S}
    C_{t,b_t^\star}(p_t^\star).
\]

Mean realized regression is
\[
    \mathrm{Regression}_{\mathrm{sel}}
    =
    \frac{
        \sum_{t\in\mathcal S}
        \sum_{b\in\mathcal B}
        m_{t,b}^{D}
        D_{t,b}(p_t^\star)
    }{
        \sum_{t\in\mathcal S}
        \sum_{b\in\mathcal B}
        m_{t,b}^{D}
    }.
\]

An edit is empirically feasible when every observed degradation
coordinate satisfies its tolerance. Let
$\mathcal B_t^D=\{b\in\mathcal B:m_{t,b}^{D}=1\}$, and define
\[
F_t(p)=\mathbb I[\mathcal B_t^D\neq\varnothing]
\prod_{b\in\mathcal B_t^D}
\mathbb I[D_{t,b}(p)\le\epsilon_b].
\]
The empirical feasibility rate is
\[
\mathrm{EFR}=\frac{1}{|\mathcal S|}
\sum_{t\in\mathcal S}F_t(p_t^\star).
\]

EFR is an empirical selection metric.
It should not be interpreted as a finite-sample certification guarantee;
formal certification is defined separately in
Appendix~\ref{app:certification_details}.

\begin{algorithm}[t]
\small
\caption{\method: State-Conditioned Rubric Optimization}
\label{alg:method}

\KwIn{
$R_0$;
$\mathcal D_{\mathrm{dyn,train}}$;
$\mathcal D_{\mathrm{dyn,val}}$;
$\mathcal D_{\mathrm{cert}}$;
rounds $T$
}
\KwOut{final specification $R_T$}

Initialize the response model from warm-start transitions\;

\For{$t=0,\ldots,T-1$}{
    Generate candidate edits $\mathcal P_t$\;

    Form the current state $s_t$\;

    Predict edit responses and uncertainty\;

    Select $p_t^\star$ using Eq.~\eqref{eq:candidate_selection}\;

    \If{a candidate is selected}{
        Measure its paired response on $\mathcal D_{\mathrm{dyn,train}}$\;

        Update the transition history and response model\;

        Compute certification bounds on held-out data\;

        \eIf{Eq.~\eqref{eq:certification_rule} holds}{
            $R_{t+1}\leftarrow R_t\oplus p_t^\star$\;
        }{
            $R_{t+1}\leftarrow R_t$\;
        }
    }
    \Else{
        $R_{t+1}\leftarrow R_t$\;
    }
}
\Return{$R_T$}\;
\end{algorithm}


\section{Prediction Variants and Ablation Controls}
\label{app:prediction_variants}

\subsection{Four Predictor Variants}

Figure~\ref{fig:ablation_triptych}(a) compares four ways of using the
same candidate edit $p$. We write $a_p=\phi(p)$ for
the text embedding of the edit and its target metadata, $h_R=\psi_R(R_t)$ for the current
rubric alone, and $h_s=\psi(R_t,\mathcal H_t,\mathbf v_t)$ for the
full state (rubric text, measured edit history, and training-answer
behavior). Every model applies the same element-wise sigmoid to its
response logits and predicts the correction/degradation coordinates
with the same masked training targets:
\[
\begin{aligned}
\eta_{\mathrm{edit}}
  &= W_a a_p+c,\\
\eta_{\mathrm{rubric}}
  &= W_a a_p+W_R h_R+c,\\
\eta_{\mathrm{history}}
  &= W_a a_p+W_s h_s+c,\\
\eta_{\mathrm{full}}
  &= W_a a_p+U[(Aa_p)\odot(Bh_s)]+c.
\end{aligned}
\]

The corresponding labels in Figure~\ref{fig:ablation_triptych}(a) are
\emph{Edit only}, \emph{Edit + current rubric}, \emph{Edit + rubric +
history}, and \emph{RubricWorld (Ours)}. The third model also reads
$\mathbf v_t$ as part of its full-state summary; ``history'' in its
short plot label therefore includes the observed grading profile.
The first three variants use additive input combinations; the full
variant learns a factorized edit--state interaction. This comparison
varies architecture while holding data splits, response masks, edit
representations, and evaluation metrics fixed.

\subsection{End-to-End Controls and Cost Accounting}

In Table~\ref{tab:world_model_e2e}, \emph{Full RubricWorld} uses the
state-conditioned predictor for candidate selection. The
\emph{w/o State} variant retains learned prediction and the same
text-embedding representation but does not condition on the rubric
state. The \emph{w/o Learned Model} variant replaces predicted effects
with directly measured paired candidate effects when selecting edits;
all variants retain the same candidate generator and independent
certification rule.

Five-round quality is measured after five refinements. The 20-round
column reports only the cumulative cost of continuing the same
per-round procedure for fifteen more rounds: grader, candidate budget,
data roles, and certification rule are unchanged. Token counts include
warm-start collection when applicable, proposal generation, paired
grading, and commit checks; final test evaluation is excluded. The
quality--cost sweep in Figure~\ref{fig:quality_cost} plots means over
three random seeds, with each point including the warm-start tokens.
Local predictor training and embedding computation are separate from
LLM-token usage and are not represented as free wall-clock compute.